\documentclass{article}

\usepackage{microtype}
\usepackage{graphicx}
\usepackage{subfigure}
\usepackage{booktabs} 

\usepackage{hyperref}

\usepackage[accepted]{icml2025}

\usepackage{amsmath}
\usepackage{amssymb}
\usepackage{mathtools}
\usepackage{amsthm}
\usepackage{paralist}

\usepackage{multirow}
\usepackage{longtable}

\usepackage[capitalize,noabbrev]{cleveref}

\theoremstyle{plain}

\theoremstyle{definition}

\theoremstyle{remark}

\usepackage{algorithmic}
\renewcommand{\algorithmicrequire}{\textbf{Input:}}
\renewcommand{\algorithmicensure}{\textbf{Output:}}
\renewcommand{\algorithmiccomment}[1]{\hfill $\triangleright$ #1}
\newcommand{\CommentLeft}[1]{\(\triangleright\) #1}

\usepackage[textsize=tiny]{todonotes}

\icmltitlerunning{PAMNet: Cycle-aware Phase-Amplitude Modulation Network for Multivariate Time Series Forecasting}

\begin{document}

\twocolumn[
\icmltitle{PAMNet: Cycle-aware Phase-Amplitude Modulation Network for Multivariate Time Series Forecasting}




\begin{icmlauthorlist}
\icmlauthor{Yingbo Zhou}{1}
\icmlauthor{Yutong Ye}{2}
\icmlauthor{Zhiwei Ling}{3}
\icmlauthor{Shuhao Li}{1}
\icmlauthor{Rui Qian}{1}
\icmlauthor{Jian Xiong}{1}
\icmlauthor{Li Sun}{4}
\icmlauthor{Dejing Dou}{1}
\end{icmlauthorlist}

\icmlaffiliation{1}{Fudan University}
\icmlaffiliation{2}{Beihang University}
\icmlaffiliation{3}{Zhejiang University}
\icmlaffiliation{4}{Beijing University of Posts and Telecommunications}

\icmlcorrespondingauthor{Dejing Dou}{doudejing@fudan.edu.cn}


\vskip 0.3in
]



\printAffiliationsAndNotice{}  

\begin{abstract}
Reliable periodic patterns serve as a fundamental basis for accurate multivariate time series forecasting. 
However, existing methods either implicitly extract periodicity through complex model architectures (e.g., Transformers) with high computational overhead or overlook the intrinsic phase-amplitude coupling when modeling periodic components explicitly.
To address these issues, we propose a novel Cycle-aware Phase-Amplitude Modulation Network (\textbf{PAMNet}) that explicitly decomposes periodic patterns into complementary phase and amplitude components.
The core innovation lies in its dual-branch modulator, featuring dedicated learnable embeddings for phase positioning and amplitude modulation.
The phase branch employs cyclical embeddings to capture phase-dependent mean shifts, while the amplitude branch models intensity variations to adapt to changes in variance. 
A lightweight modulator with element-wise fusion efficiently combines these components, enabling explicit modeling of their interactions without complex attention mechanisms. 
Extensive experiments on twelve real-world datasets demonstrate that our method achieves state-of-the-art performance through its novel phase-amplitude decoupling mechanism, offering a new perspective for cyclical modeling in time series forecasting.
\end{abstract}

\section{Introduction}
\label{intro}

Multivariate Time Series (MTS) forecasting represents a critical task with widespread applications across numerous domains, including energy management \cite{DBLP:conf/icaai/VolkovsUC23}, financial analysis \cite{DBLP:conf/cikm/ZhuLHLCL24}, traffic planning \cite{DBLP:journals/tits/ShuCX22}, and healthcare monitoring \cite{DBLP:journals/tmis/MoridSD23}. 
In many of these real-world scenarios, time series data exhibit inherent periodic patterns, ranging from daily and weekly cycles in electricity consumption to seasonal variations in economic indicators, which serve as fundamental anchors for reliable predictions. 
Effectively capturing and leveraging these periodic structures remains a central challenge in advancing time series forecasting capabilities.

Current approaches to modeling periodicity in MTS forecasting can be broadly categorized into two paradigms. 
The first paradigm employs sophisticated neural architectures, particularly Transformers \cite{DBLP:conf/nips/WuXWL21, DBLP:conf/aaai/ZhouZPZLXZ21, DBLP:conf/iclr/NieNSK23} and their variants \cite{DBLP:conf/icml/ZhouMWW0022, DBLP:conf/iclr/ZhangY23, DBLP:conf/icml/YuZ0A0W24}, to implicitly extract periodic patterns through attention mechanisms and positional encodings. 
While these methods have shown impressive performance, they suffer from substantial computational complexity and limited interpretability into how periodicity is actually captured and utilized. 
The second paradigm explicitly models periodic components through seasonal-trend decomposition \cite{DBLP:conf/nips/LiuZCXLM022, DBLP:conf/aaai/ZengCZ023, DBLP:conf/iclr/WuHLZ0L23, DBLP:conf/iclr/WangWSHLMZ024, DBLP:conf/iclr/0001WLLB0X24} or learnable cycle structure \cite{DBLP:conf/icml/Lin0WCY24, DBLP:conf/nips/Lin0HWMZ24, lin2025TQNet}, but typically overlooks the crucial coupling between phase and amplitude—the precise timing of periodic events and their intensity variations. 
Such periodic non-stationarity is particularly problematic in real-world data, due to external factors such as holidays, weather events, or market disruptions.

\begin{figure*}[htbp]
\begin{center}
\includegraphics[width=0.98\textwidth]{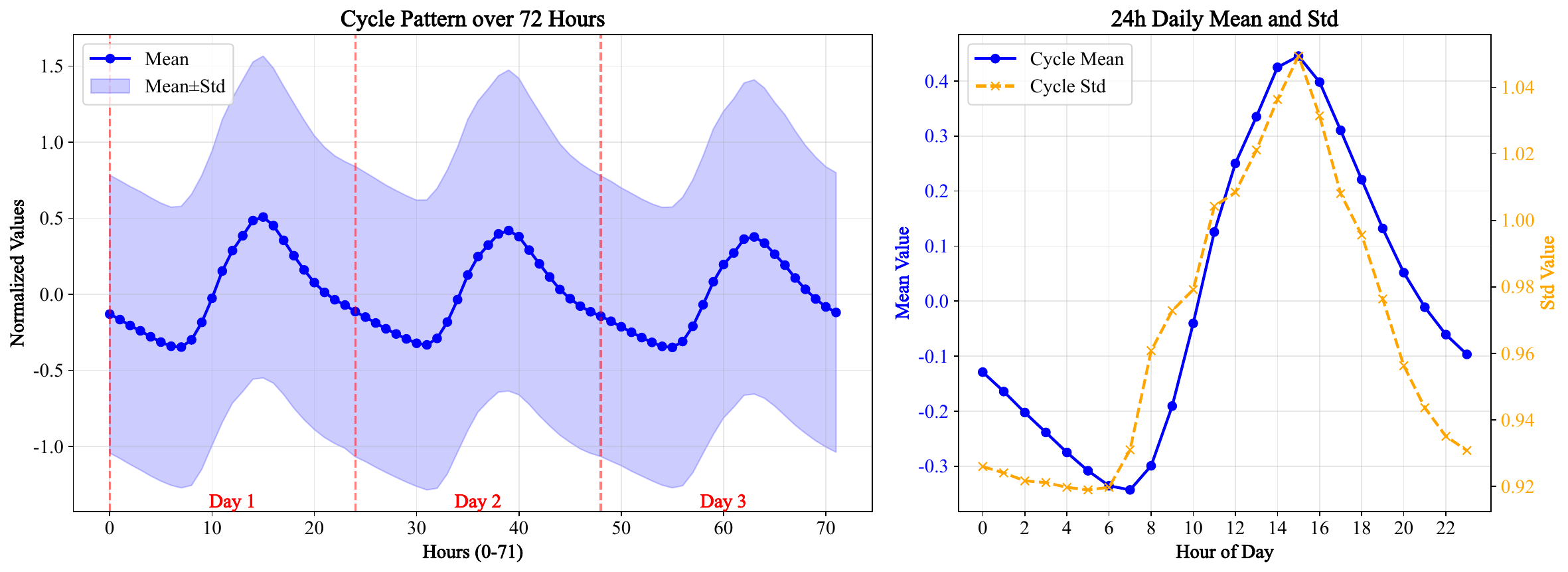}
\vspace{-0.3cm}
\caption{Periodic non-stationarity in time series data. The evolution of the cycle mean and cycle standard deviation for channel OT in ETTh1 reveals a consistent periodic pattern in its statistical properties, confirming periodic non-stationarity rather than random drift.}
\label{fig:1}
\vspace{-0.4cm}
\end{center}
\end{figure*}

To precisely diagnose this limitation of overlooking phase-amplitude coupling, we provide a pilot visualization on a typical industrial dataset (ETTh1) in Figure \ref{fig:1}.  
While the mean curve of the left subfigure confirms stable phase alignment across three consecutive daily cycles, the fluctuating $\pm\text{Std}$ band demonstrates significant amplitude variations at identical phase positions.
More importantly, the synchronous oscillation of mean (phase) and standard deviation (amplitude) in the right subfigure indicates that temporal positioning and intensity modulation are intrinsically linked yet independently varying. 
Such phase-amplitude coupling demonstrates that treating periodicity as a monolithic component is fundamentally limited, necessitating an explicit decomposition approach for effective modeling.

Driven by this observation, we propose a novel Cycle-aware Phase-Amplitude Modulation Network (\textbf{PAMNet}) that explicitly decomposes periodic patterns into complementary phase and amplitude components.
Its core innovation lies in a dual-stream modulation that structurally mirrors the phase-amplitude duality, moving beyond monolithic periodicity modeling. 
Without heavy attention mechanisms, we implement a lightweight phase-amplitude decoupling framework by an efficient feature-wise modulation strategy, explicitly modeling the interactions between the phase and amplitude streams.
Comprehensive empirical studies on multiple real-world benchmarks demonstrate that PAMNet achieves state-of-the-art forecasting accuracy through its novel phase-amplitude modulation paradigm.

In summary, our main contributions are threefold:
\begin{compactitem}
    \item We formalize the critical role of phase-amplitude decoupling as a more effective paradigm for modeling periodicity in time series, moving beyond the prevailing monolithic or implicitly decomposed approaches.
    \item We propose a novel cycle-aware phase-amplitude modulation network, PAMNet, which features dedicated cyclical embeddings and a lightweight modulator for effective and explicit decoupling interactions.
    \item Extensive empirical evidence shows that our method achieves state-of-the-art performance across several domains, which provides a more expressive framework for cyclical pattern modeling.
\end{compactitem}

\section{Related Work}
\label{related_work}


\textbf{Implicit Periodic Modeling in Time Series Forecasting.}
The evolution of deep learning for time series forecasting has been characterized by increasingly sophisticated architectures designed to implicitly capture temporal dependencies, including periodic patterns. 
Early approaches based on Recurrent Neural Networks \cite{DBLP:conf/nips/RangapuramSGSWJ18, DBLP:journals/corr/abs-2308-11200} and Temporal Convolutional Networks \cite{DBLP:conf/nips/LiuZCXLM022, DBLP:conf/iclr/LuoW24} established foundational capabilities for sequence modeling, though their ability to explicitly represent long-range periodic structures remained limited. 
The recent dominance of Transformer-based architectures, including FEDformer \cite{DBLP:conf/icml/ZhouMWW0022}, iTransformer \cite{DBLP:conf/iclr/LiuHZWWML24}, and other variants \cite{lin2025TQNet, liu2025timebridge}, has further advanced this implicit paradigm through specialized attention mechanisms or frequency-domain integrations that subsume periodic patterns within their general-purpose learning frameworks. 
While achieving impressive performance, these methods inherently suffer from substantial computational costs, limited interpretability in how periodic patterns are represented, and sensitivity to distribution shifts in periodic dynamics — largely due to the lack of strong inductive biases for cyclic structures.
In contrast, we design a lightweight architecture with a built-in structural prior for periodicity, thereby inherently enhancing both efficiency and interpretability.

\begin{figure*}[htbp]
\begin{center}
\includegraphics[width=0.98\textwidth]{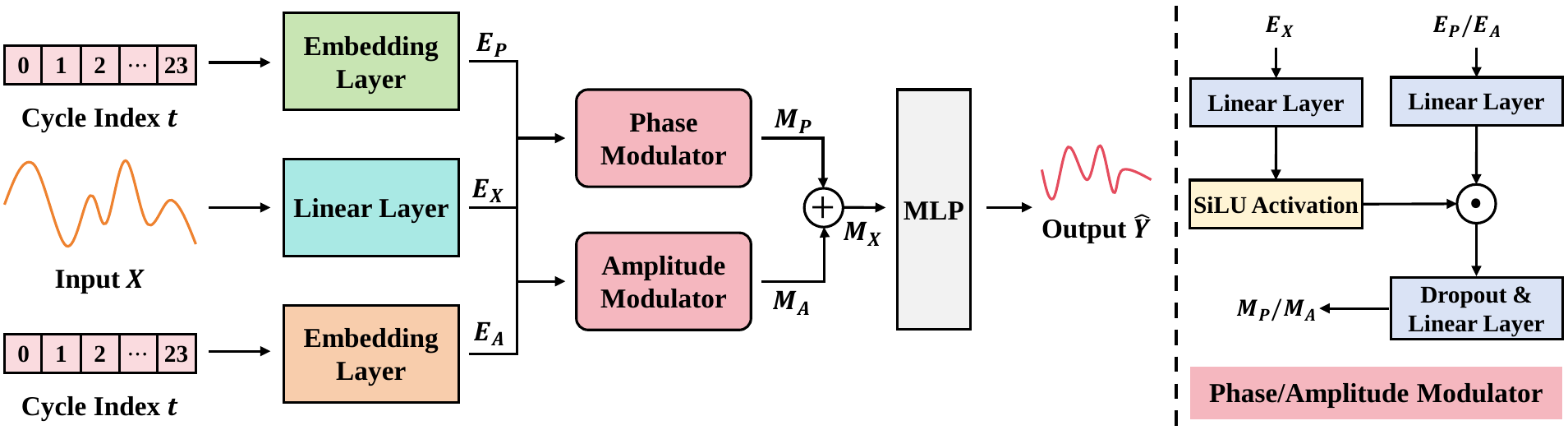}
\vspace{-0.3cm}
\caption{Overview of PAMNet, a novel framework inspired by signal modulation. The architecture begins by generating adaptive phase $E_P$ and amplitude $E_A$ carrier signals from a cycle index $t$. These carriers then modulate the encoded time series $E_X$ within two parallel paths, which explicitly capture cyclical dynamics through a phase-amplitude decoupling modulator. The resulting representations are fused and passed through a final MLP to produce the forecasting values $\hat{Y}$.}
\label{fig:2}
\vspace{-0.4cm}
\end{center}
\end{figure*}

\textbf{Explicit Periodic Modeling in Time Series Forecasting.}
In contrast to implicit methods, a distinct research trajectory has focused on explicitly incorporating periodic structures through various structural priors. 
This paradigm shares a common philosophy of treating periodicity as an architecturally explicit component, either through predefined structural assumptions or learned periodic representations, such as employing seasonal-trend decomposition \cite{DBLP:conf/aaai/ZengCZ023, DBLP:conf/iclr/WangWSHLMZ024, DBLP:conf/nips/DengYY0T024}, temporal transformation \cite{DBLP:conf/iclr/WuHLZ0L23, DBLP:conf/icml/Lin0WCY24}, and learnable cycle representations \cite{DBLP:conf/nips/Lin0HWMZ24, lin2025TQNet}. 
However, these methods predominantly model periodicity at a macroscopic level, representing it as a monolithic seasonal component or fixed basis functions, which fails to disentangle the two fundamental, interdependent dimensions of periodic patterns: the stable phase (temporal positioning within cycles) and the dynamic amplitude (intensity variation across cycles). 
Consequently, they cannot adequately capture the intricate phase-amplitude coupling phenomenon essential for modeling structural periodic patterns. 
In contrast, our approach pioneers a fine-grained, explicit decoupling of temporal patterns into complementary phase and amplitude components, directly enabling a more nuanced and robust modeling of cyclical dynamics.

\textbf{Relationship to Concurrent Work}. 
A concurrently submitted work, PAMod \cite{ConcurrentWork1}, also employs the terminology of phase-amplitude modulation. 
However, our PAMNet is fundamentally different in both objective and mechanism.
PAMod adapts to modeling cyclical non-stationarity via distribution correction, while PAMNet actively disentangles and reconstructs the cyclical signal itself.

\section{Methodology}

Multivariate time series forecasting aims to predict future values of a system based on its historical observations.
Formally, given a historical sequence of $L$ time steps with $N$ variates, denoted as $X=\{x_1, x_2, \cdots, x_L\}\in\mathbb{R}^{L\times N}$, the goal is to learn a mapping $F(\cdot)$ to the subsequent $H$ time steps $\{X_{L+1}, X_{L+2}, \cdots, X_{L+H}\}\in\mathbb{R}^{H\times N}$.
A predominant challenge in real-world forecasting is non-stationary periodicity \cite{DBLP:journals/dsp/JavorskyjYMZM17}, where the underlying cyclical patterns (e.g., daily, weekly) exhibit significant variations in both phase (e.g., shifting peak hours) and amplitude (e.g., fluctuating intensity) over time. 
To address this, we draw inspiration from signal processing and learn to modulate the historical base signal with adaptive carriers that explicitly model evolving cyclical dynamics.

\subsection{Structure Overview}

We propose \textbf{PAMNet} (Phase-Amplitude Modulation Network), a novel framework that conceptualizes time series forecasting as a process of adaptive signal modulation. 
As illustrated in Figure \ref{fig:2}, PAMNet is architected around two core learnable components:
\begin{compactitem}
    \item \textbf{Dual Carrier Embeddings}: A cyclical index $t$, representing a position within a presumed period (e.g., hour-of-day), is fed into two separate embedding layers to generate a phase carrier $E_P(t)$ and an amplitude carrier $E_A(t)$.
    \item \textbf{Parallel Modulation Pathways}: The encoded historical time series $E_X$ is processed in two parallel streams. Each stream is modulated with a specialized phase and amplitude modulator module, producing a phase-modulated representation $M_P$ and an amplitude-modulated representation $M_A$.
\end{compactitem}
These two complementary representations are fused and passed through an output projection MLP to generate the final prediction $\hat{Y}$. This design enables an explicit and decoupled modeling of phase and amplitude variations in non-stationary cycles.

\subsection{Key Components of PAMNet}

\textbf{Input Encoding and Cyclical Conditioning.} 
The raw input $X$ is first projected into a latent representation $E_X\in\mathbb{R}^{N\times d}$ using an inverted data embedding layer \cite{DBLP:conf/iclr/LiuHZWWML24}, which transposes the time points as variates tokens:
\begin{equation}
    E_X=X^{\top}W+b~\in\mathbb{R}^{N\times d}, 
\end{equation}
where $d$ is the embedding dimension. 
Concurrently, a crucial component of our framework is the cyclical index $t$.
Inspired by \cite{DBLP:conf/nips/Lin0HWMZ24}, the index is derived from the absolute time stamp corresponding to the last observed time point in the input sequence $X$.
Specifically, let $\tau$ be the absolute time index of the sequence's end. 
The cyclical index $t$ is then computed as the remainder of $\tau_{end}$ divided by the predefined cycle length $c$:
\begin{equation}
    t=\tau_{end}~\text{mod}~c~~\in[0, c-1].
\end{equation}
The scalar integer $t$ indicates the temporal position within a cycle and serves as the key to adapting our model to non-stationary periodicity. 
It conditions the model on the specific phase of the cycle at the forecast origin, enabling it to generate context-aware predictions.

\textbf{Learnable Carrier Embeddings.}
Leveraging the cyclical index $t$ defined above, our model generates adaptive carrier signals through dedicated embedding layers. 
This design allows the model to discover optimal, data-specific carrier waveforms, moving beyond the limitations of pre-defined sinusoidal functions, as shown in our ablation study of Section \ref{sec:4.3}.
The phase carrier $E_P(t)$ is obtained via a lookup from the learnable phase embedding matrix $\Omega_p\in\mathbb{R}^{c\times d}$:
\begin{equation}
    E_P(t)=\text{Lookup}(\Omega_p, t)~~\in\mathbb{R}^{N\times d}.
\end{equation}
The carrier $E_P(t)$ is designed to encode information relevant to the temporal alignment of patterns within a cycle.
The amplitude carrier $E_A(t)$ is obtained via another lookup from the learnable amplitude embedding matrix $\Omega_a\in\mathbb{R}^{c\times N\cdot d}$ and then reshaped to align with the variate and feature dimensions:
\begin{equation}
    E_A(t)=\text{Reshape}(\text{Lookup}(\Omega_a, t))~~\in \mathbb{R}^{N\times d}.
\end{equation}
The carrier $E_A(t)$ is tailored to encode information about the relative intensity or prominence of temporal patterns across different variates.

\textbf{Phase and Amplitude Modulator.}
The core operation of our model is performed by the Modulator module, which implements the generalized modulation process.
For a given carrier signal $S$ (which can be $E_P(t)$ or $E_A(t)$) and the baseband signal $E_X$, the Modulator $\mathcal{M}$ is defined as:
\begin{equation}
    \mathcal{M}(E_X, S)=W_3(\text{Dropout}(\text{SiLU}(W_1E_X)\odot W_2S)),
\end{equation}
where $W_1, W_2, W_3$ are learnable linear projections, $\text{SiLU}(\cdot)$ is the Sigmoid Linear Unit activation function \cite{DBLP:journals/nn/ElfwingUD18}, and $\odot$ denotes the Hadamard (element-wise) product. This module is instantiated in two parallel paths (i.e., phase modulation $M_P$ and amplitude modulation $M_A$), each modulated by a semantically distinct carrier:
\begin{equation}
    \begin{aligned}
        M_P&=\mathcal{M}(E_X, E_P(t)),\\
        M_A&=\mathcal{M}(E_X, E_A(t)).
    \end{aligned}
\end{equation}
$M_P$ modulates the input using the phase carrier to align temporal contexts while $M_A$ modulates the input using the amplitude carrier to scale feature intensities.

The element-wise product $\odot$ serves as the fundamental modulation operator. It allows each carrier to dynamically and non-linearly transform the baseband representation $E_X$ by performing feature-wise gating. The Modulator thus actively shapes the temporal representation based on the contextual information encoded within the carrier, which is pivotal for capturing non-stationary dynamics.

\textbf{Output Fusion and Prediction.}
The modulated representations from both paths are fused by simple summation $M_X=M_P+M_A$.
The fused representation integrates both phase-aligned and amplitude-scaled information. 
Finally, a lightweight Multi-Layer Perceptron (MLP) projects $M_X$ to the forecasting horizon:
\begin{equation}
    \hat{Y}=\text{MLP}(M_X).
\end{equation}

\subsection{Theoretical Analysis in a Modulation Perspective}

Our framework is grounded in the mathematical principles of signal modulation. 
While the model employs two distinct pathways labeled ``phase" and ``amplitude", their underlying operation is unified under the formalism of Amplitude Modulation (AM) \cite{carlson1968communication}.

In classical AM, a baseband signal $m(t)$ is multiplied by a carrier wave $c(t)$: 
\begin{equation}
    s_{AM}(t)=m(t)\cdot c(t).
\end{equation}
Our modulator module is a direct implementation of this operation in a high-dimensional feature space, where $\text{SiLU}(W_1E_X)$ is the projected baseband signal and $W_2S$ is the projected carrier.
The innovation of PAMNet lies in its use of two specialized, learnable carriers:
\begin{compactitem}
    \item The phase carrier $E_P(t)$ is optimized to perform a feature-wise phase alignment via AM. The element-wise product $\odot$ effectively ``rotates" or shifts the feature activations, encoding the temporal context (``when") within the cycle.
    \item The amplitude carrier $E_A(t)$ is optimized to perform a feature-wise intensity scaling via AM. The same operation $\odot$ here amplifies or attenuates specific features, encoding the magnitude context (``how much").
\end{compactitem}
Inspired by signal processing, PAMNet conceptualizes its operation through the lens of Phase and Amplitude Modulation. 
However, in our neural network realization, both modulation types are implemented through a unified, learnable amplitude modulation mechanism in a high-dimensional embedding space. 
The semantic distinction between ``phase" and ``amplitude" modulation arises not from the mathematical operation (both use Hadamard product), but from the specialized roles learned by the respective carriers, $E_P(t)$ and $E_A(t)$. 
The former guides temporal alignment while the latter governs intensity scaling, thereby enabling the explicit modeling of the decoupled evolution in mean (phase) and standard deviation (amplitude), which characterize the periodic non-stationarity illustrated in Figure \ref{fig:1}.
We provide more interpretable insights in Section \ref{sec:4.5}.

\begin{table*}[htbp]
\centering
\caption{Multivariate forecasting performance. The lookback length is set to $L=96$ and all the results are averaged from all predictions $H\in\{12, 24, 48, 96\}$ for PEMS and $H\in\{96, 192, 336, 720\}$ for other benchmarks. See Table \ref{tab:7} in the Appendix for the full results.}
\vspace{0.2cm}
\label{tab:1}
\scriptsize
\setlength{\tabcolsep}{1.2mm}{
\begin{tabular}{c|cc|cc|cc|cc|cc|cc|cc|cc|cc|cc}
\specialrule{0.15em}{0pt}{2pt}
\multirow{2}{*}{Model} & \multicolumn{2}{c|}{\textbf{PAMNet}}     & \multicolumn{2}{c|}{TQNet}   & \multicolumn{2}{c|}{FilterTS} & \multicolumn{2}{c|}{Amplifier} & \multicolumn{2}{c|}{CycleNet} & \multicolumn{2}{c|}{TimeMixer} & \multicolumn{2}{c|}{iTransformer} & \multicolumn{2}{c|}{PatchTST} & \multicolumn{2}{c|}{TimesNet} & \multicolumn{2}{c}{DLinear} \\ 
\cmidrule(lr){2-3} \cmidrule(lr){4-5} \cmidrule(lr){6-7} \cmidrule(lr){8-9} \cmidrule(lr){10-11} \cmidrule(lr){12-13} \cmidrule(lr){14-15} \cmidrule(lr){16-17} \cmidrule(lr){18-19} \cmidrule(lr){20-21}
                       & \multicolumn{2}{c|}{(\textbf{Ours})}     & \multicolumn{2}{c|}{(\citeyear{lin2025TQNet})}    & \multicolumn{2}{c|}{(\citeyear{DBLP:conf/aaai/WangLDW25})}     & \multicolumn{2}{c|}{(\citeyear{DBLP:conf/aaai/Fei000N25})}      & \multicolumn{2}{c|}{(\citeyear{DBLP:conf/nips/Lin0HWMZ24})}     & \multicolumn{2}{c|}{(\citeyear{DBLP:conf/iclr/WangWSHLMZ024})}      & \multicolumn{2}{c|}{(\citeyear{DBLP:conf/iclr/LiuHZWWML24})}         & \multicolumn{2}{c|}{(\citeyear{DBLP:conf/iclr/NieNSK23})}     & \multicolumn{2}{c|}{(\citeyear{DBLP:conf/iclr/WuHLZ0L23})}     & \multicolumn{2}{c}{(\citeyear{DBLP:conf/aaai/ZengCZ023})}    \\ 
                       \specialrule{0.10em}{1pt}{1pt}
Metric                 & MSE            & MAE            & MSE            & MAE         & MSE           & MAE           & MSE            & MAE           & MSE           & MAE           & MSE             & MAE          & MSE               & MAE           & MSE           & MAE           & MSE           & MAE           & MSE          & MAE          \\ 
\specialrule{0.10em}{1pt}{1pt}
ETTm1                  & {\color{red}\textbf{0.365}} & {\color{red}\textbf{0.384}} & {\color{blue}\underline{0.377}}    & {\color{blue}\underline{0.393}} & 0.385         & 0.396         & 0.382          & 0.395         & 0.379         & 0.396         & 0.381           & 0.395        & 0.407             & 0.410         & 0.387         & 0.400         & 0.400         & 0.406         & 0.403        & 0.407        \\ 
\specialrule{0.10em}{1pt}{1pt}
ETTm2                  & {\color{red}\textbf{0.264}} & {\color{red}\textbf{0.308}} & 0.277          & 0.323       & 0.277         & 0.322         & 0.280          & 0.326         & {\color{blue}\underline{0.266}}   & {\color{blue}\underline{0.314}}   & 0.275           & 0.323        & 0.288             & 0.332         & 0.281         & 0.326         & 0.291         & 0.333         & 0.350        & 0.401        \\ 
\specialrule{0.10em}{1pt}{1pt}
ETTh1                  & {\color{red}\textbf{0.420}} & {\color{red}\textbf{0.424}} & 0.441          & 0.434       & 0.434         & 0.430         & {\color{blue}\underline{0.430}}    & {\color{blue}\underline{0.428}}   & 0.457         & 0.441         & 0.447           & 0.440        & 0.454             & 0.448         & 0.469         & 0.455         & 0.458         & 0.450         & 0.456        & 0.452        \\ 
\specialrule{0.10em}{1pt}{1pt}
ETTh2                  & {\color{blue}\underline{0.374}}    & {\color{red}\textbf{0.393}} & 0.378          & 0.402       & 0.375         & 0.398         & 0.381          & 0.405         & 0.388         & 0.409         & {\color{red}\textbf{0.364}}  & {\color{blue}\underline{0.395}}  & 0.383             & 0.407         & 0.387         & 0.407         & 0.414         & 0.427         & 0.559        & 0.515        \\ 
\specialrule{0.10em}{1pt}{1pt}
ECL                    & {\color{red}\textbf{0.161}} & {\color{red}\textbf{0.251}} & {\color{blue}\underline{0.164}}    & {\color{blue}\underline{0.259}}       & 0.180         & 0.272         & 0.172          & 0.266         & 0.168         & {\color{blue}\underline{0.259}}         & 0.182           & 0.272        & 0.178             & 0.270         & 0.205         & 0.290         & 0.193         & 0.295         & 0.212        & 0.300        \\ 
\specialrule{0.10em}{1pt}{1pt}
Traffic                & {\color{blue}\underline{0.440}}    & {\color{red}\textbf{0.263}} & 0.445          & {\color{blue}\underline{0.276}} & 0.470         & 0.315         & 0.483          & 0.317         & 0.472         & 0.314         & 0.484           & 0.297        & {\color{red}\textbf{0.428}}    & 0.282         & 0.481         & 0.300         & 0.620         & 0.336         & 0.625        & 0.383        \\ 
\specialrule{0.10em}{1pt}{1pt}
Weather                & {\color{red}\textbf{0.240}} & {\color{red}\textbf{0.263}} & {\color{blue}\underline{0.242}}    & {\color{blue}\underline{0.269}} & 0.245         & 0.274         & 0.253          & 0.275         & 0.243         & 0.271         & 0.240           & 0.271        & 0.258             & 0.278         & 0.259         & 0.273         & 0.259         & 0.287         & 0.265        & 0.317        \\ 
\specialrule{0.10em}{1pt}{1pt}
Solar                  & {\color{blue}\underline{0.204}}    & {\color{red}\textbf{0.228}} & {\color{red}\textbf{0.198}} & {\color{blue}\underline{0.256}} & 0.215         & 0.277         & 0.241          & 0.270         & 0.210         & 0.261         & 0.216           & 0.280        & 0.233             & 0.262         & 0.270         & 0.307         & 0.301         & 0.319         & 0.330        & 0.401        \\ 
\specialrule{0.10em}{1pt}{1pt}
PEMS03                 & {\color{red}\textbf{0.091}} & {\color{red}\textbf{0.188}} & {\color{blue}\underline{0.097}}    & {\color{blue}\underline{0.203}} & 0.134         & 0.246         & 0.131          & 0.239         & 0.118         & 0.226         & 0.167           & 0.267        & 0.113             & 0.222         & 0.180         & 0.291         & 0.147         & 0.248         & 0.278        & 0.375        \\ 
\specialrule{0.10em}{1pt}{1pt}
PEMS04                 & {\color{red}\textbf{0.085}} & {\color{red}\textbf{0.180}} & {\color{blue}\underline{0.091}}    & {\color{blue}\underline{0.197}} & 0.125         & 0.241         & 0.135          & 0.249         & 0.119         & 0.232         & 0.185           & 0.287        & 0.111             & 0.221         & 0.195         & 0.307         & 0.129         & 0.241         & 0.295        & 0.388        \\ 
\specialrule{0.10em}{1pt}{1pt}
PEMS07                 & {\color{blue}\underline{0.078}}    & {\color{red}\textbf{0.163}} & {\color{red}\textbf{0.075}} & {\color{blue}\underline{0.171}} & 0.120         & 0.220         & 0.122          & 0.226         & 0.113         & 0.214         & 0.181           & 0.271        & 0.101             & 0.204         & 0.211         & 0.303         & 0.125         & 0.226         & 0.329        & 0.396        \\ 
\specialrule{0.10em}{1pt}{1pt}
PEMS08                 & {\color{red}\textbf{0.137}} & {\color{red}\textbf{0.211}} & {\color{blue}\underline{0.142}}    & 0.229       & 0.180         & 0.266         & 0.183          & 0.271         & 0.150         & 0.246         & 0.226           & 0.299        & 0.150             & {\color{blue}\underline{0.226}}   & 0.280         & 0.321         & 0.193         & 0.271         & 0.379        & 0.416        \\ 
\specialrule{0.10em}{1pt}{1pt}
{$1^{\text{st}}$ Count}         & {\color{red}\bf{8}}         & \multicolumn{1}{c|}{\color{red}\bf{12}}         &  {\color{blue}\underline{2}}     & \multicolumn{1}{c|}{0}      & 0       & \multicolumn{1}{c|}{0}      & 0      & \multicolumn{1}{c|}{0}      & 0      & \multicolumn{1}{c|}{0}      & 1      & \multicolumn{1}{c|}{0}  & 1   &\multicolumn{1}{c|}{0}      &  {0}     & \multicolumn{1}{c|}{0}   &  0     & \multicolumn{1}{c|}{0}   & 0 & 0      \\ 
\specialrule{0.15em}{2pt}{0pt}
\end{tabular}}
\vspace{-0.3cm}
\end{table*}

\section{Experiments}
To validate the effectiveness of the proposed PAMNet, we conduct extensive experiments on several real-world datasets covering both long-term and short-term forecasting tasks in four prediction horizons. 
Our experiments are designed to answer the following key questions:
\begin{compactitem}
    \item \textbf{Q1 (Superiority)}: How does PAMNet perform against state-of-the-art methods?
    \item \textbf{Q2 (Effectiveness)}: How does each core component contribute to the model's capability?
    \item \textbf{Q3 (Generalization)}: How does PAMNet generalize to different forecasting scenarios?
    \item \textbf{Q4: (Interpretability)}: How does the model's internal process provide interpretable insights?
\end{compactitem}

\vspace{-0.2cm}
\subsection{Experimental Settings}
\vspace{-0.1cm}
\textbf{Datasets.}
We evaluate the proposed method on twelve widely used real-world datasets. 
Spanning various domains, these include the ETT series \cite{DBLP:conf/aaai/ZhouZPZLXZ21}, PEMS series \cite{DBLP:conf/nips/LiuZCXLM022}, as well as Electricity, Weather, Traffic, and Solar datasets \cite{DBLP:conf/nips/WuXWL21}. 
Following the mainstream setting \cite{DBLP:conf/iclr/LiuHZWWML24, lin2025TQNet},  the prediction horizons are set to $\{12, 24, 48, 96\}$ for the PEMS series, and $\{96, 192, 336, 720\}$ for the rest benchmarks. 
This selection ensures coverage across a broad spectrum of data properties, including scale, dimensionality, and domain-specific temporal patterns.

\textbf{Baselines.} 
To rigorously evaluate the performance of PAMNet, we compare it with nine recent and representative models,  including TQNet \cite{lin2025TQNet}, FilterTS \cite{DBLP:conf/aaai/WangLDW25}, Amplifier \cite{DBLP:conf/aaai/Fei000N25}, CycleNet \cite{DBLP:conf/nips/Lin0HWMZ24}, TimeMixer \cite{DBLP:conf/iclr/WangWSHLMZ024}, iTransformer \cite{DBLP:conf/iclr/LiuHZWWML24}, TimesNet \cite{DBLP:conf/iclr/WuHLZ0L23}, PatchTST \cite{DBLP:conf/iclr/NieNSK23}, and DLinear \cite{DBLP:conf/aaai/ZengCZ023}.
In adherence to standard practice for a fair comparison, all models are trained with a consistent look-back length of 96.

\textbf{Implementation Details.} 
All experiments are implemented in PyTorch \cite{DBLP:conf/nips/PaszkeGMLBCKLGA19} and trained on a single NVIDIA GeForce RTX 4090 GPU (24GB) using the Adam optimizer \cite{DBLP:journals/corr/KingmaB14}. 
Following prior work \cite{DBLP:conf/iclr/0049PS0YY0L0T25}, we employ a hybrid MAE loss that operates in both the time and frequency domains to ensure stable training. 
The key hyperparameter $c$, which defines the cycle length, is configured for each dataset according to the guidelines established in CycleNet and TQNet.
Finally, we use the standard Mean Squared Error (MSE) and Mean Absolute Error (MAE) for performance evaluation.

\vspace{-0.1cm}
\subsection{Main Performance (Q1)}
To answer Q1 regarding PAMNet's superiority, we conduct a comprehensive comparison in both predictive effectiveness and computational efficiency.

\textbf{Predictive Effectiveness.}
As shown in Table \ref{tab:1}, PAMNet achieves Top-1 performance in 20 out of 24 forecasting metrics across 12 datasets, establishing new state-of-the-art accuracy. 
This superiority stems from two key advantages: (1) Compared to models that implicitly capture periodicity (e.g., iTransformer, PatchTST), our explicit cyclical modeling provides more consistent gains on strongly periodic data; (2) Even against recent explicit cyclical approaches like TQNet and CycleNet, PAMNet demonstrates clear improvements—our decoupled phase-amplitude modulation offers a more flexible inductive bias, better adapting to real-world non-stationary and multi-scale periodic variations.

\begin{figure*}[htbp]
\begin{center}
\includegraphics[width=0.98\textwidth]{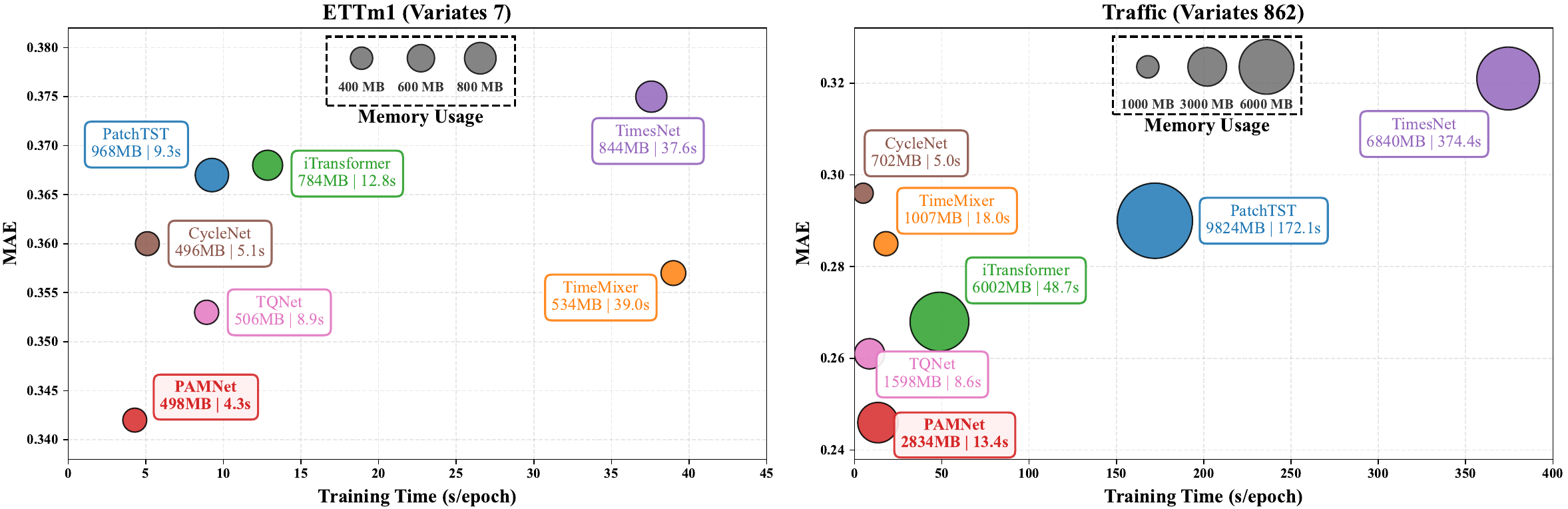}
\vspace{-0.3cm}
\caption{Model efficiency comparison under the input-96-predict-96 setting. All results are obtained by using the official model configurations with the same batch sizes: 32 and 16 for the ETTm1 and Traffic datasets, respectively.}
\label{fig:3}
\vspace{-0.4cm}
\end{center}
\end{figure*}

\begin{table*}[!h]
\centering
\caption{Ablation study of PAMNet. The input length $L$ and forecasting horizon $H$ are 96.}
\label{tab:2}
\vspace{0.15cm}
\scriptsize
\setlength{\tabcolsep}{0.8mm}{
\begin{tabular}{c|cc|cccccccccccccccc|cccc}
\specialrule{0.15em}{0pt}{2pt}
Variants                 & \multicolumn{2}{c|}{PAMNet}     & \multicolumn{16}{c|}{Modulator}                                                                                                                                                                                                                                                                                                                                                                                                                                                              & \multicolumn{4}{c}{Training Loss}                                                                                                                                                  \\ 
\specialrule{0.15em}{2pt}{2pt}
                           & \multicolumn{2}{c|}{SiLU}       & \multicolumn{2}{c|}{Tanh}                                          & \multicolumn{2}{c|}{Sigmoid}                                                                       & \multicolumn{2}{c|}{ReLU}                                                                          & \multicolumn{2}{c|}{GELU}                                          & \multicolumn{2}{c|}{w/o $E_A$}      & \multicolumn{2}{c|}{w/o $E_P$}      & \multicolumn{2}{c|}{w Sinusoidal}  & \multicolumn{2}{c|}{w/o Modulator} & \multicolumn{2}{c|}{MSE Loss}                                                                      & \multicolumn{2}{c}{MAE Loss}                                                  \\ 
                           \cmidrule(lr){2-3} \cmidrule(lr){4-5} \cmidrule(lr){6-7} \cmidrule(lr){8-9} \cmidrule(lr){10-11} \cmidrule(lr){12-13} \cmidrule(lr){14-15} \cmidrule(lr){16-17} \cmidrule(lr){18-19} \cmidrule(lr){20-21} \cmidrule(lr){22-23}
\multirow{-2}{*}{Datasets} & MSE            & MAE            & MSE                                   & \multicolumn{1}{c|}{MAE}   & MSE                                   & \multicolumn{1}{c|}{MAE}                                   & MSE                                   & \multicolumn{1}{c|}{MAE}                                   & MSE                                   & \multicolumn{1}{c|}{MAE}   & MSE   & \multicolumn{1}{c|}{MAE}   & MSE   & \multicolumn{1}{c|}{MAE}   & MSE   & \multicolumn{1}{c|}{MAE}   & MSE              & MAE             & MSE                                   & \multicolumn{1}{c|}{MAE}                                   & MSE                                   & MAE                                   \\ 
\specialrule{0.10em}{1pt}{1pt}
ETTm1                      & \textbf{0.308} & \textbf{0.347} & 0.311                                 & \multicolumn{1}{c|}{0.348} & {\color{red} \textbf{0.300}} & \multicolumn{1}{c|}{{\color{red} \textbf{0.340}}} & {\color{red} \textbf{0.305}} & \multicolumn{1}{c|}{{\color{red} \textbf{0.346}}} & {\color{red} \textbf{0.307}} & \multicolumn{1}{c|}{0.347} & 0.315 & \multicolumn{1}{c|}{0.354} & 0.312 & \multicolumn{1}{c|}{0.350} & 0.310 & \multicolumn{1}{c|}{0.348} & 0.322            & 0.368           & {\color{red} \textbf{0.305}} & \multicolumn{1}{c|}{{\color{red} \textbf{0.340}}} & {\color{red} \textbf{0.303}} & {\color{red} \textbf{0.338}} \\ 
\specialrule{0.10em}{1pt}{1pt}
ETTm2                      & \textbf{0.165} & \textbf{0.243} & 0.166                                 & \multicolumn{1}{c|}{0.243} & 0.165                                 & \multicolumn{1}{c|}{{\color{red} \textbf{0.242}}} & {\color{red} \textbf{0.163}} & \multicolumn{1}{c|}{{\color{red} \textbf{0.242}}} & 0.165                                 & \multicolumn{1}{c|}{0.244} & 0.170 & \multicolumn{1}{c|}{0.248} & 0.168 & \multicolumn{1}{c|}{0.246} & 0.167 & \multicolumn{1}{c|}{0.244} & 0.173            & 0.252           & 0.168                                 & \multicolumn{1}{c|}{0.245}                                 & 0.165                                 & 0.243                                 \\ 
\specialrule{0.10em}{1pt}{1pt}
ETTh1                      & \textbf{0.357} & \textbf{0.383} & 0.358                                 & \multicolumn{1}{c|}{0.385} & 0.359                                 & \multicolumn{1}{c|}{0.386}                                 & {\color{red} \textbf{0.355}} & \multicolumn{1}{c|}{{\color{red} \textbf{0.382}}} & 0.356                                 & \multicolumn{1}{c|}{0.383} & 0.367 & \multicolumn{1}{c|}{0.389} & 0.364 & \multicolumn{1}{c|}{0.386} & 0.363 & \multicolumn{1}{c|}{0.387} & 0.374            & 0.389           & 0.367                                 & \multicolumn{1}{c|}{0.385}                                 & 0.367                                 & 0.384                                 \\ 
\specialrule{0.10em}{1pt}{1pt}
ETTh2                      & \textbf{0.285} & \textbf{0.330} & 0.287                                 & \multicolumn{1}{c|}{0.331} & {\color{red} \textbf{0.282}} & \multicolumn{1}{c|}{0.330}                                 & 0.287                                 & \multicolumn{1}{c|}{0.331}                                 & 0.290                                 & \multicolumn{1}{c|}{0.332} & 0.292 & \multicolumn{1}{c|}{0.335} & 0.290 & \multicolumn{1}{c|}{0.333} & 0.288 & \multicolumn{1}{c|}{0.332} & 0.295            & 0.340           & 0.290                                 & \multicolumn{1}{c|}{0.332}                                 & 0.289                                 & 0.331                                 \\ 
\specialrule{0.10em}{1pt}{1pt}
ECL                        & \textbf{0.132} & \textbf{0.222} & {\color{red} \textbf{0.131}} & \multicolumn{1}{c|}{0.223} & {\color{red} \textbf{0.130}} & \multicolumn{1}{c|}{{\color{red} \textbf{0.221}}} & {\color{red} \textbf{0.131}} & \multicolumn{1}{c|}{0.222}                                 & {\color{red} \textbf{0.131}} & \multicolumn{1}{c|}{0.222} & 0.149 & \multicolumn{1}{c|}{0.235} & 0.138 & \multicolumn{1}{c|}{0.228} & 0.145 & \multicolumn{1}{c|}{0.232} & 0.177            & 0.253           & 0.136                                 & \multicolumn{1}{c|}{0.225}                                 & 0.134                                 & 0.222                                 \\ 
\specialrule{0.10em}{1pt}{1pt}
Traffic                    & \textbf{0.415} & \textbf{0.246} & {\color{red} \textbf{0.414}} & \multicolumn{1}{c|}{0.246} & {\color{red} \textbf{0.413}} & \multicolumn{1}{c|}{0.247}                                 & 0.415                                 & \multicolumn{1}{c|}{0.248}                                 & 0.420                                 & \multicolumn{1}{c|}{0.248} & 0.427 & \multicolumn{1}{c|}{0.255} & 0.424 & \multicolumn{1}{c|}{0.250} & 0.425 & \multicolumn{1}{c|}{0.253} & 0.476            & 0.296           & 0.424                                 & \multicolumn{1}{c|}{0.245}                                 & 0.424                                 & 0.239                                 \\ 
\specialrule{0.10em}{1pt}{1pt}
Weather                    & \textbf{0.154} & \textbf{0.193} & 0.156                                 & \multicolumn{1}{c|}{0.195} & 0.154                                 & \multicolumn{1}{c|}{{\color{red} \textbf{0.192}}} & 0.155                                 & \multicolumn{1}{c|}{0.193}                                 & 0.156                                 & \multicolumn{1}{c|}{0.196} & 0.172 & \multicolumn{1}{c|}{0.205} & 0.162 & \multicolumn{1}{c|}{0.199} & 0.171 & \multicolumn{1}{c|}{0.206} & 0.184            & 0.217           & 0.158                                 & \multicolumn{1}{c|}{0.199}                                 & 0.156                                 & 0.193                                 \\ 
\specialrule{0.10em}{1pt}{1pt}
Solar                      & \textbf{0.189} & \textbf{0.214} & 0.194                                 & \multicolumn{1}{c|}{0.217} & 0.192                                 & \multicolumn{1}{c|}{0.215}                                 & 0.192                                 & \multicolumn{1}{c|}{0.218}                                 & 0.192                                 & \multicolumn{1}{c|}{0.219} & 0.199 & \multicolumn{1}{c|}{0.226} & 0.195 & \multicolumn{1}{c|}{0.223} & 0.193 & \multicolumn{1}{c|}{0.226} & 0.220            & 0.259           & {\color{red} \textbf{0.186}} & \multicolumn{1}{c|}{0.223}                                 & 0.193                                 & 0.212                                 \\ 
\specialrule{0.10em}{1pt}{1pt}
PEMS03                     & \textbf{0.134} & \textbf{0.225} & 0.136                                 & \multicolumn{1}{c|}{0.226} & 0.139                                 & \multicolumn{1}{c|}{0.228}                                 & {\color{red} \textbf{0.133}} & \multicolumn{1}{c|}{0.225}                                 & 0.135                                 & \multicolumn{1}{c|}{0.225} & 0.175 & \multicolumn{1}{c|}{0.272} & 0.146 & \multicolumn{1}{c|}{0.243} & 0.167 & \multicolumn{1}{c|}{0.265} & 0.231            & 0.328           & 0.135                                 & \multicolumn{1}{c|}{0.227}                                 & {\color{red} \textbf{0.133}} & {\color{red} \textbf{0.223}} \\ 
\specialrule{0.10em}{1pt}{1pt}
PEMS04                     & \textbf{0.106} & \textbf{0.201} & {\color{red} \textbf{0.104}} & \multicolumn{1}{c|}{0.200} & {\color{red} \textbf{0.104}} & \multicolumn{1}{c|}{0.201}                                 & {\color{red} \textbf{0.105}} & \multicolumn{1}{c|}{0.202}                                 & 0.107                                 & \multicolumn{1}{c|}{0.202} & 0.177 & \multicolumn{1}{c|}{0.277} & 0.118 & \multicolumn{1}{c|}{0.190} & 0.170 & \multicolumn{1}{c|}{0.268} & 0.259            & 0.346           & 0.109                                 & \multicolumn{1}{c|}{0.211}                                 & 0.107                                 & 0.202                                 \\ 
\specialrule{0.10em}{1pt}{1pt}
PEMS07                     & \textbf{0.109} & \textbf{0.187} & 0.119                                 & \multicolumn{1}{c|}{0.188} & 0.117                                 & \multicolumn{1}{c|}{0.189}                                 & {\color{red} \textbf{0.107}} & \multicolumn{1}{c|}{0.187}                                 & 0.109                                 & \multicolumn{1}{c|}{0.187} & 0.152 & \multicolumn{1}{c|}{0.247} & 0.123 & \multicolumn{1}{c|}{0.209} & 0.156 & \multicolumn{1}{c|}{0.248} & 0.281            & 0.348           & 0.110                                 & \multicolumn{1}{c|}{0.193}                                 & 0.109                                 & 0.188                                 \\ 
\specialrule{0.10em}{1pt}{1pt}
PEMS08                     & \textbf{0.239} & \textbf{0.267} & 0.252                                 & \multicolumn{1}{c|}{0.270} & 0.253                                 & \multicolumn{1}{c|}{0.273}                                 & {\color{red} \textbf{0.228}} & \multicolumn{1}{c|}{0.273}                                 & {\color{red} \textbf{0.236}} & \multicolumn{1}{c|}{0.267} & 0.306 & \multicolumn{1}{c|}{0.335} & 0.251 & \multicolumn{1}{c|}{0.272} & 0.293 & \multicolumn{1}{c|}{0.333} & 0.595            & 0.505           & 0.243                                 & \multicolumn{1}{c|}{0.270}                                 & 0.240                                 & 0.268                                 \\ 
\specialrule{0.10em}{1pt}{1pt}
Avg                        & \textbf{0.216} & \textbf{0.255} & 0.219                                 & \multicolumn{1}{c|}{0.256} & 0.217                                 & \multicolumn{1}{c|}{0.255}                                 & {\color{red} \textbf{0.215}} & \multicolumn{1}{c|}{0.256}                                 & 0.217                                 & \multicolumn{1}{c|}{0.256} & 0.242 & \multicolumn{1}{c|}{0.282} & 0.224 & \multicolumn{1}{c|}{0.261} & 0.237 & \multicolumn{1}{c|}{0.279} & 0.299            & 0.325           & 0.219                                 & \multicolumn{1}{c|}{0.258}                                 & 0.218                                 & 0.254                                 \\ 
\specialrule{0.15em}{2pt}{0pt}
\end{tabular}}
\vspace{-0.25cm}
\end{table*}

\textbf{Computational Efficiency.}
Beyond accuracy, we comprehensively compare the forecasting performance, training speed, and memory usage of PAMNet with other baselines, using the official model configurations and the same batch size. 
As shown in Figure \ref{fig:3}, PAMNet demonstrates competitive and often superior computational efficiency compared to contemporary baselines, establishing a favorable trade-off between performance and resource utilization. 
On ETTm1, PAMNet attains a memory footprint on par with CycleNet while delivering faster training iterations, and it notably outperforms TQNet in both memory and speed. 
On Traffic, although outperformed in efficiency by specialized cyclical models (CycleNet and TQNet), PAMNet still maintains a clear advantage over widely-used complex architectures such as iTransformer, PatchTST, and TimesNet.
This efficiency comes from the lightweight dual-path modulation design of PAMNet, which retains expressiveness through explicit linear-complexity phase-amplitude disentanglement and avoids the quadratic cost of global self-attention.
Thus, PAMNet achieves an effective balance between computational efficiency and state-of-the-art accuracy, offering a practical solution for forecasting tasks requiring joint optimization of both criteria.

\subsection{Ablation Studies (Q2)}
\label{sec:4.3}

We systematically ablate the core designs of PAMNet to validate their necessity and robustness, focusing on architectural components and hyperparameter cycle length $c$.

\textbf{Architectural Components.}
We ablate key design choices of PAMNet to analyze their impact in Table \ref{tab:2}.
(1) \textbf{Activation Functions}: replacing the default SiLU with Tanh, Sigmoid, ReLU, or GELU leads to only marginal variations in average MSE (0.216\textendash 0.219) and MAE (0.255\textendash 0.256), indicating that PAMNet's performance is robust to the choice of nonlinearity. 
(2) \textbf{Optimization Objective}: using pure MSE or MAE loss yields nearly identical results to the default hybrid MAE loss, confirming that accuracy gains stem primarily from the architecture design rather than the optimization objective. 
(3) \textbf{Phase-Amplitude Modulator}: removing the dual-path modulator causes the most severe performance drop (Avg. MSE: $0.216 \rightarrow 0.299$), especially on strongly periodic datasets such as Traffic (MSE: $0.415 \rightarrow 0.476$); 
Ablating either the phase embedding $E_P$ or the amplitude embedding $E_A$ also leads to clear degradation, underscoring that disentangling phase and amplitude is essential for capturing complex cyclical dynamics.
Replacing the learnable phase-amplitude embeddings with a fixed sinusoidal function leads to a noticeable performance drop (Avg. MSE: 0.216 $\rightarrow$ 0.237), confirming that real-world cyclical patterns are often non-stationary and cannot be fully captured by predefined periodic bases. 

\begin{figure}[htbp]
\begin{center}
\includegraphics[width=0.98\linewidth]{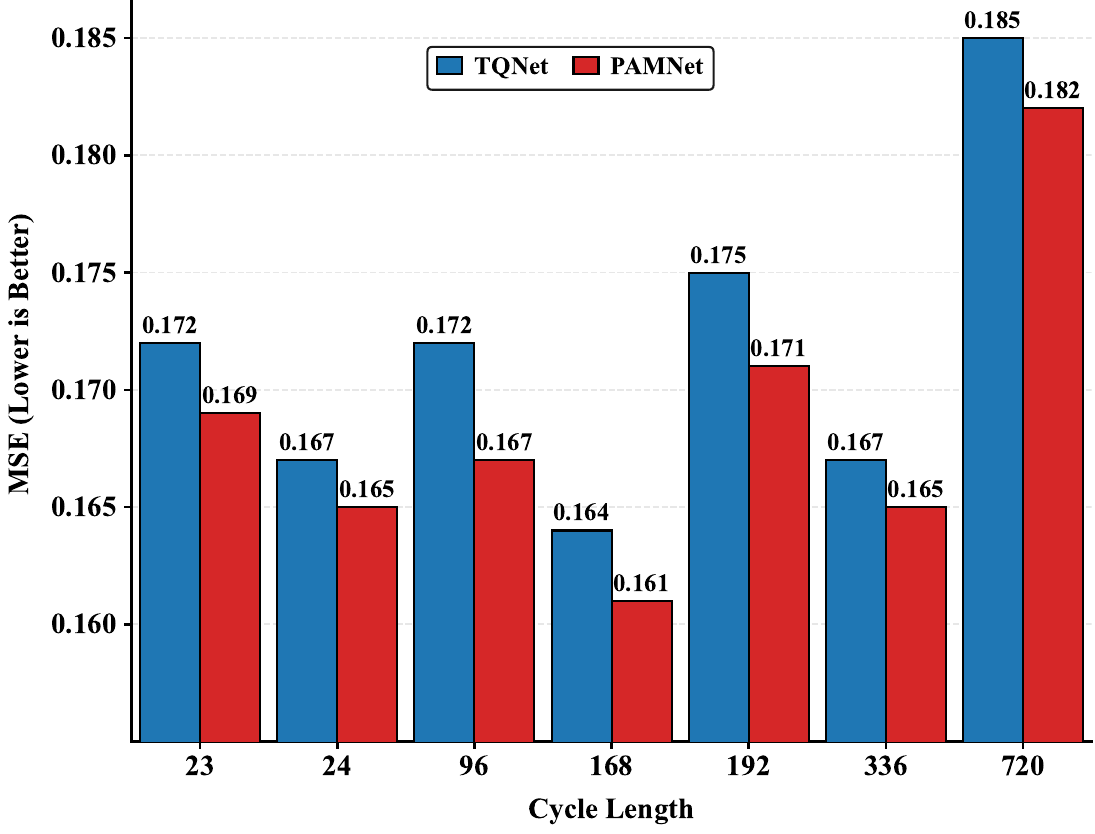}
\vspace{-0.3cm}
\caption{Ablation performance (MSE) on the ECL dataset with varying cycle length $c$. All results are averaged across forecasting horizons $H\in\{96, 192, 336, 720\}$.}
\label{fig:4}
\vspace{-0.4cm}
\end{center}
\end{figure}

\textbf{Cycle Length Sensitivity.}
We systematically vary the hyperparameter cycle length $c=\{23, 24, 96, 168, 192, 336, 720\}$ to assess its impact on both PAMNet and a leading cyclical baseline, TQNet. 
As shown in \ref{fig:4}, optimal performance for both models occurs near $c=168$, aligning with natural periodicities in the data, while overly small or large $c$ leads to degradation. 
Notably, PAMNet exhibits greater robustness to variations in $c$, where its MSE varies within a narrower interval (0.161\textendash 0.182) compared to TQNet (0.164\textendash 0.185). 
These results demonstrate that while both models benefit from a well-specified cycle length, PAMNet’s phase-amplitude decoupling mechanism makes it less sensitive to suboptimal choices—a valuable property in practical settings where the true periodicity may not be known precisely.

\subsection{Generalization Analysis (Q3)}
\label{sec:4.4}

\begin{figure}[htbp]
\begin{center}
\includegraphics[width=0.98\linewidth]{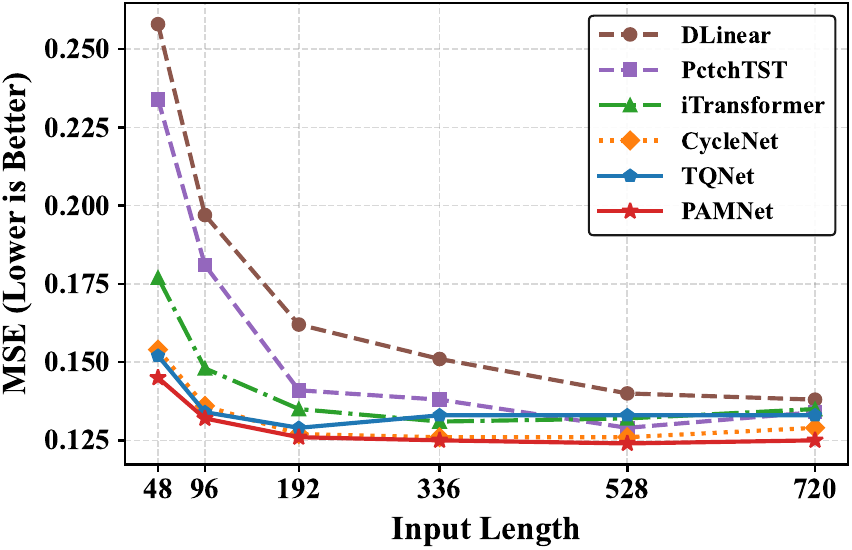}
\vspace{-0.3cm}
\caption{Impact of varying lookback lengths on ECL.}
\label{fig:5}
\vspace{-0.4cm}
\end{center}
\end{figure}

We further assess PAMNet’s adaptability across the following three generalization scenarios. 

\textbf{Impact of Input Length.}
As shown in Figure \ref{fig:5}, We vary the look-back window from 48 to 720 while keeping the prediction horizon fixed at 96.
PAMNet consistently achieves the lowest MSE at every length, outperforming the second-best model by 0.8\%\textendash 4.6\%.
Crucially, PAMNet exhibits the strongest robustness to length variation, where its MSE fluctuates only by 0.021 across the entire range, far less than other models (e.g., DLinear: 0.120).
This demonstrates that the phase-amplitude modulation effectively distills temporal dependencies even from short sequences, delivering stable and superior accuracy under varying history inputs.

\vspace{-0.35cm}
\begin{table}[htbp]
\centering
\caption{Performance under missing values on ECL.}
\label{tab:3}
\vspace{0.15cm}
\scriptsize
\setlength{\tabcolsep}{0.9mm}{
\begin{tabular}{cc|cc|cccc|cccc}
\specialrule{0.15em}{0pt}{2pt}
\multicolumn{2}{c|}{\multirow{2}{*}{ECL}}                   & \multicolumn{2}{c|}{\multirow{2}{*}{Standard}} & \multicolumn{4}{c|}{Zeros}                                                             & \multicolumn{4}{c}{Noise}                                                              \\ 
\cmidrule(lr){5-8}\cmidrule(lr){9-12}
\multicolumn{2}{c|}{}                                       & \multicolumn{2}{c|}{}                          & \multicolumn{2}{c|}{$p=0.1$}                             & \multicolumn{2}{c|}{$p=0.3$}        & \multicolumn{2}{c|}{$p=0.1$}                             & \multicolumn{2}{c}{$p=0.3$}         \\ 
\specialrule{0.10em}{1pt}{1pt}
\multicolumn{2}{c|}{Metric}                                 & MSE                    & MAE                   & MSE            & \multicolumn{1}{c|}{MAE}            & MSE            & MAE            & MSE            & \multicolumn{1}{c|}{MAE}            & MSE            & MAE            \\ 
\specialrule{0.10em}{1pt}{1pt}
\multicolumn{1}{c|}{\multirow{5}{*}{\rotatebox{90}{iTransformer}}}    & 96  & 0.148                  & 0.240                 & 0.157          & \multicolumn{1}{c|}{0.253}          & 0.195          & 0.298          & 0.157          & \multicolumn{1}{c|}{0.254}          & 0.193          & 0.295          \\
\multicolumn{1}{c|}{}                                 & 192 & 0.162                  & 0.253                 & 0.171          & \multicolumn{1}{c|}{0.265}          & 0.208          & 0.308          & 0.171          & \multicolumn{1}{c|}{0.267}          & 0.203          & 0.305          \\
\multicolumn{1}{c|}{}                                 & 336 & 0.178                  & 0.269                 & 0.183          & \multicolumn{1}{c|}{0.278}          & 0.221          & 0.319          & 0.190          & \multicolumn{1}{c|}{0.289}          & 0.221          & 0.325          \\
\multicolumn{1}{c|}{}                                 & 720 & 0.225                  & 0.317                 & 0.228          & \multicolumn{1}{c|}{0.319}          & 0.256          & 0.344          & 0.227          & \multicolumn{1}{c|}{0.322}          & 0.262          & 0.355          \\ 
\cmidrule(lr){2-12}
\multicolumn{1}{c|}{}                                 & Avg & 0.178                  & 0.270                 & 0.185          & \multicolumn{1}{c|}{0.279}          & 0.220          & 0.317          & 0.186          & \multicolumn{1}{c|}{0.283}          & 0.220          & 0.320          \\ 
\specialrule{0.10em}{1pt}{1pt}
\multicolumn{1}{c|}{\multirow{5}{*}{\rotatebox{90}{TimeMixer}}}       & 96  & 0.153                  & 0.247                 & 0.155          & \multicolumn{1}{c|}{0.259}          & 0.252          & 0.370          & 0.157          & \multicolumn{1}{c|}{0.261}          & 0.250          & 0.354          \\
\multicolumn{1}{c|}{}                                 & 192 & 0.166                  & 0.256                 & 0.172          & \multicolumn{1}{c|}{0.273}          & 0.240          & 0.346          & 0.170          & \multicolumn{1}{c|}{0.271}          & 0.278          & 0.375          \\
\multicolumn{1}{c|}{}                                 & 336 & 0.185                  & 0.277                 & 0.189          & \multicolumn{1}{c|}{0.290}          & 0.327          & 0.420          & 0.191          & \multicolumn{1}{c|}{0.293}          & 0.237          & 0.342          \\
\multicolumn{1}{c|}{}                                 & 720 & 0.225                  & 0.310                 & 0.234          & \multicolumn{1}{c|}{0.325}          & 0.323          & 0.416          & 0.236          & \multicolumn{1}{c|}{0.328}          & 0.279          & 0.375          \\ 
\cmidrule(lr){2-12}
\multicolumn{1}{c|}{}                                 & Avg & 0.182                  & 0.272                 & 0.188          & \multicolumn{1}{c|}{0.287}          & 0.286          & 0.388          & 0.189          & \multicolumn{1}{c|}{0.288}          & 0.261          & 0.362          \\ 
\specialrule{0.10em}{1pt}{1pt}
\multicolumn{1}{c|}{\multirow{5}{*}{\rotatebox{90}{TQNet}}}           & 96  & 0.134                  & 0.229                 & 0.138          & \multicolumn{1}{c|}{0.235}          & 0.155          & 0.257          & 0.137          & \multicolumn{1}{c|}{0.233}          & 0.156          & 0.256          \\
\multicolumn{1}{c|}{}                                 & 192 & 0.154                  & 0.247                 & 0.156          & \multicolumn{1}{c|}{0.254}          & 0.172          & 0.271          & 0.158          & \multicolumn{1}{c|}{0.255}          & 0.176          & 0.274          \\
\multicolumn{1}{c|}{}                                 & 336 & 0.169                  & 0.264                 & 0.172          & \multicolumn{1}{c|}{0.269}          & 0.189          & 0.289          & 0.174          & \multicolumn{1}{c|}{0.269}          & 0.196          & 0.294          \\
\multicolumn{1}{c|}{}                                 & 720 & 0.201                  & 0.294                 & 0.214          & \multicolumn{1}{c|}{0.306}          & 0.235          & 0.328          & 0.210          & \multicolumn{1}{c|}{0.301}          & 0.242          & 0.333          \\ 
\cmidrule(lr){2-12}
\multicolumn{1}{c|}{}                                 & Avg & 0.164                  & 0.259                 & 0.170          & \multicolumn{1}{c|}{0.266}          & 0.188          & 0.286          & 0.170          & \multicolumn{1}{c|}{0.265}          & 0.193          & 0.289          \\ 
\specialrule{0.10em}{1pt}{1pt}
\multicolumn{1}{c|}{\multirow{5}{*}{\textbf{\rotatebox{90}{PAMNet}}}} & 96  & \textbf{0.132}         & \textbf{0.222}        & \textbf{0.134} & \multicolumn{1}{c|}{\textbf{0.225}} & \textbf{0.145} & \textbf{0.241} & \textbf{0.135} & \multicolumn{1}{c|}{\textbf{0.227}} & \textbf{0.145} & \textbf{0.240} \\
\multicolumn{1}{c|}{}                                 & 192 & \textbf{0.148}         & \textbf{0.238}        & \textbf{0.151} & \multicolumn{1}{c|}{\textbf{0.241}} & \textbf{0.164} & \textbf{0.258} & \textbf{0.152} & \multicolumn{1}{c|}{\textbf{0.243}} & \textbf{0.163} & \textbf{0.257} \\
\multicolumn{1}{c|}{}                                 & 336 & \textbf{0.165}         & \textbf{0.256}        & \textbf{0.167} & \multicolumn{1}{c|}{\textbf{0.258}} & \textbf{0.180} & \textbf{0.275} & \textbf{0.169} & \multicolumn{1}{c|}{\textbf{0.261}} & \textbf{0.179} & \textbf{0.273} \\
\multicolumn{1}{c|}{}                                 & 720 & \textbf{0.199}         & \textbf{0.287}        & \textbf{0.202} & \multicolumn{1}{c|}{\textbf{0.290}} & \textbf{0.220} & \textbf{0.311} & \textbf{0.205} & \multicolumn{1}{c|}{\textbf{0.293}} & \textbf{0.219} & \textbf{0.307} \\ 
\cmidrule(lr){2-12}
\multicolumn{1}{c|}{}                                 & Avg & \textbf{0.161}         & \textbf{0.251}        & \textbf{0.164} & \multicolumn{1}{c|}{\textbf{0.254}} & \textbf{0.177} & \textbf{0.271} & \textbf{0.165} & \multicolumn{1}{c|}{\textbf{0.256}} & \textbf{0.177} & \textbf{0.269} \\ 
\specialrule{0.15em}{2pt}{0pt}
\end{tabular}}
\end{table}

\textbf{Performance under Missing Values.}
In real-world scenarios, time-series data often contain missing values due to sensor failures or transmission errors. 
To evaluate PAMNet's robustness under such conditions, we simulate incomplete historical data by randomly masking input timesteps with probability $p\in\{0.1, 0.3\}$. Masked positions are replaced with either zeros or Gaussian noise $\mathcal{N}(0, \sigma^2)$, where $\sigma$ is the empirical standard deviation of the input sequence. 
The model is trained with these corrupted inputs but evaluated on the original complete test sets.
As shown in Table \ref{tab:3}, PAMNet achieve exceptional robustness when faced with incomplete data, outperforming all baselines under both zero-imputation and noise-corruption scenarios. With 30\% missing values, PAMNet's average MSE increases by only 9.94\% (to 0.177), significantly less than TQNet (+14.63\%, to 0.188) and iTransformer (+23.6\%, to 0.220). 
Such advantage comes from its phase-amplitude disentanglement mechanism, which preserves global cyclical patterns despite local corruptions—enabling effective filtering of random noise and reliable reconstruction of zero-masked segments. 

\vspace{-0.35cm}
\begin{table}[htbp]
\centering
\caption{Performance in multivariate-to-univariate forecasting task.}
\label{tab:4}
\vspace{0.10cm}
\scriptsize
\setlength{\tabcolsep}{0.9mm}{
\begin{tabular}{cc|cc|cc|cc|cc|cc}
\specialrule{0.15em}{0pt}{2pt}
\multicolumn{2}{c|}{\multirow{2}{*}{Models}}        & \multicolumn{2}{c|}{\textbf{PAMNet}}     & \multicolumn{2}{c|}{TQNet} & \multicolumn{2}{c|}{CrossLinear} & \multicolumn{2}{c|}{TimeXer} & \multicolumn{2}{c}{iTransformer} \\ 
\cmidrule(lr){3-4}\cmidrule(lr){5-6}\cmidrule(lr){7-8}\cmidrule(lr){9-10}\cmidrule(lr){11-12}
\multicolumn{2}{c|}{}                               & \multicolumn{2}{c|}{(\textbf{Ours})}       & \multicolumn{2}{c|}{(\citeyear{lin2025TQNet})}  & \multicolumn{2}{c|}{(\citeyear{DBLP:conf/kdd/ZhouLL0025})}        & \multicolumn{2}{c|}{(\citeyear{DBLP:conf/nips/WangWDQZLQWL24})}    & \multicolumn{2}{c}{(\citeyear{DBLP:conf/iclr/LiuHZWWML24})}         \\ 
\specialrule{0.10em}{1pt}{1pt}
\multicolumn{2}{c|}{Metric}                         & MSE            & MAE            & MSE              & MAE     & MSE             & MAE            & MSE           & MAE          & MSE             & MAE            \\ 
\specialrule{0.10em}{1pt}{1pt}
\multicolumn{1}{c|}{\multirow{5}{*}{\rotatebox{90}{ETTm2}}}   & 96  & \textbf{0.060} & \textbf{0.176} & 0.064            & 0.181   & 0.064           & 0.180          & 0.067         & 0.180        & 0.071           & 0.194          \\
\multicolumn{1}{c|}{}                         & 192 & \textbf{0.092} & \textbf{0.225} & 0.097            & 0.231   & 0.098           & 0.232          & 0.101         & 0.236        & 0.108           & 0.247          \\
\multicolumn{1}{c|}{}                         & 336 & \textbf{0.121} & \textbf{0.269} & 0.128            & 0.272   & 0.128           & 0.271          & 0.130         & 0.275        & 0.140           & 0.288          \\
\multicolumn{1}{c|}{}                         & 720 & \textbf{0.174} & \textbf{0.321} & 0.181            & 0.330   & 0.183           & 0.332          & 0.182         & 0.332        & 0.188           & 0.340          \\ 
\cmidrule(lr){2-12}
\multicolumn{1}{c|}{}                         & Avg & \textbf{0.112} & \textbf{0.248} & 0.118            & 0.254   & 0.118           & 0.254          & 0.120         & 0.258        & 0.127           & 0.267          \\ 
\specialrule{0.10em}{1pt}{1pt}
\multicolumn{1}{c|}{\multirow{5}{*}{\rotatebox{90}{ECL}}}     & 96  & \textbf{0.232} & \textbf{0.341} & 0.239            & 0.348   & 0.251           & 0.359          & 0.261         & 0.366        & 0.299           & 0.403          \\
\multicolumn{1}{c|}{}                         & 192 & \textbf{0.277} & \textbf{0.369} & 0.283            & 0.375   & 0.294           & 0.381          & 0.316         & 0.397        & 0.321           & 0.413          \\
\multicolumn{1}{c|}{}                         & 336 & 0.345          & \textbf{0.411} & \textbf{0.342}   & 0.415   & 0.343           & 0.416          & 0.367         & 0.429        & 0.379           & 0.446          \\
\multicolumn{1}{c|}{}                         & 720 & 0.413 & \textbf{0.455} & 0.427            & 0.477   & \textbf{0.403}           & 0.465          & 0.365         & 0.439        & 0.461           & 0.504          \\ 
\cmidrule(lr){2-12}
\multicolumn{1}{c|}{}                         & Avg & \textbf{0.317} & \textbf{0.394} & 0.323            & 0.404   & 0.323           & 0.405          & 0.327         & 0.408        & 0.365           & 0.442          \\ 
\specialrule{0.10em}{1pt}{1pt}
\multicolumn{1}{c|}{\multirow{5}{*}{\rotatebox{90}{Traffic}}} & 96  & 0.132 & \textbf{0.193} & \textbf{0.129}            & 0.201   & 0.149           & 0.223          & 0.151         & 0.224        & 0.156           & 0.236          \\
\multicolumn{1}{c|}{}                         & 192 & \textbf{0.125} & \textbf{0.196} & 0.131            & 0.204   & 0.149           & 0.225          & 0.152         & 0.229        & 0.156           & 0.237          \\
\multicolumn{1}{c|}{}                         & 336 & \textbf{0.127} & \textbf{0.201} & 0.131            & 0.208   & 0.148           & 0.229          & 0.150         & 0.232        & 0.154           & 0.243          \\
\multicolumn{1}{c|}{}                         & 720 & \textbf{0.142} & \textbf{0.215} & 0.148            & 0.228   & 0.161           & 0.247          & 0.172         & 0.253        & 0.177           & 0.268          \\ 
\cmidrule(lr){2-12}
\multicolumn{1}{c|}{}                         & Avg & \textbf{0.132} & \textbf{0.201} & 0.135            & 0.210   & 0.152           & 0.231          & 0.156         & 0.235        & 0.161           & 0.246          \\ 
\specialrule{0.15em}{2pt}{0pt}
\end{tabular}}
\end{table}

\textbf{Forecasting with Exogenous Variables.} 
Following the setting of \cite{DBLP:conf/kdd/ZhouLL0025, DBLP:conf/nips/WangWDQZLQWL24}, we evaluate PAMNet's cross-variable generalization in the multivariate-to-univariate forecasting task.
In this setting, the model is trained on both the target series and exogenous covariates but evaluated solely on the target series. 
As shown in Table \ref{tab:4}, PAMNet consistently outperforms recent baselines, including TQNet, CrossLinear, and TimeXer, across the ETTm2, ECL, and Traffic datasets.
On average, PAMNet achieves 2.0\% \textendash 4.3\% lower MSE and MAE than the strongest competitor (TQNet).
Such improvements demonstrate that PAMNet's phase-amplitude modulation effectively extracts and integrates informative patterns from exogenous variables, enhancing the accuracy of target series forecasting.

\subsection{Interpretable Cases (Q4)}
\label{sec:4.5}

\begin{figure}[htbp]
\begin{center}
\includegraphics[width=1.0\linewidth]{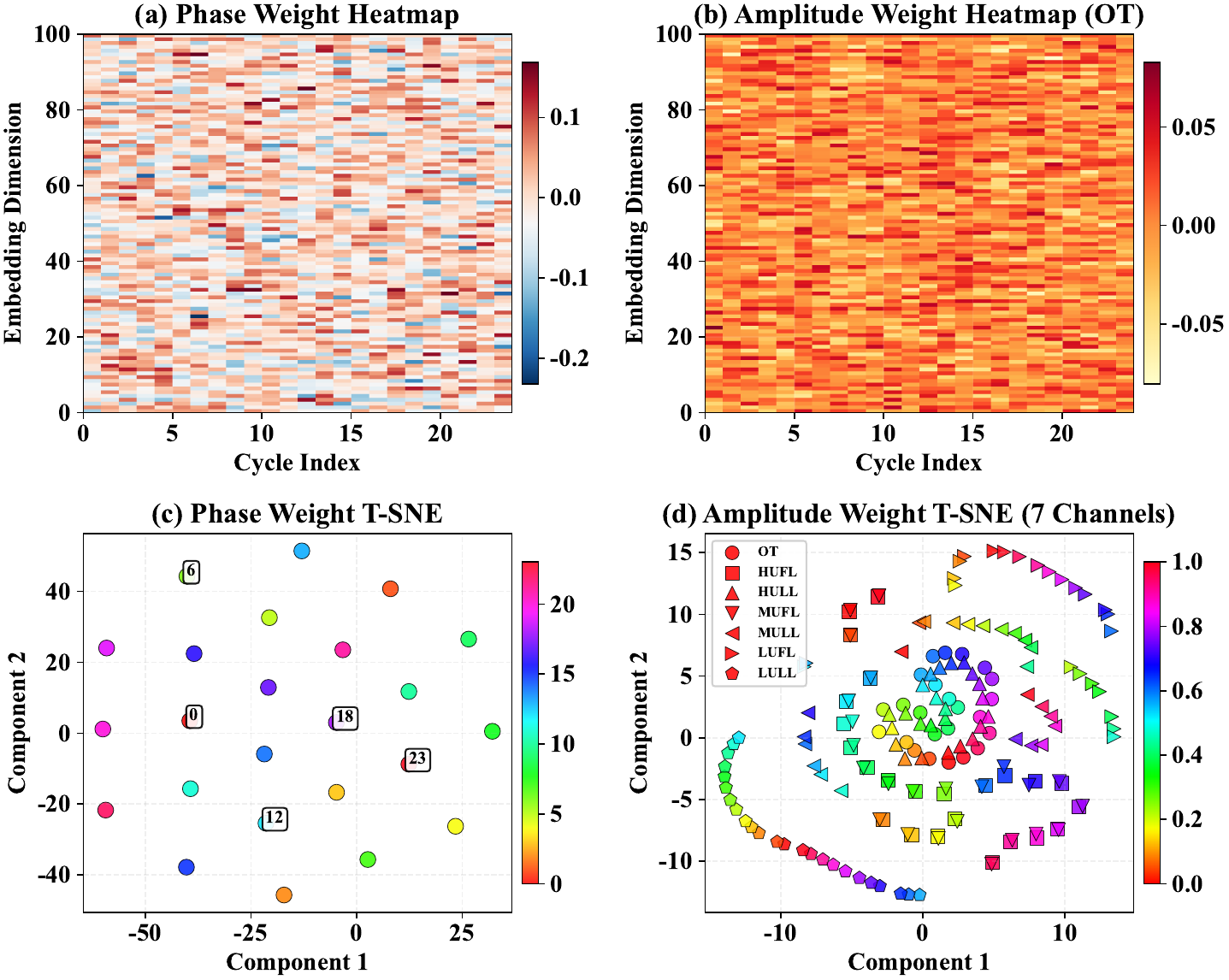}
\vspace{-0.5cm}
\caption{The learned phase and amplitude weights on ETTh1.}
\label{fig:6}
\vspace{-0.4cm}
\end{center}
\end{figure}

\begin{figure}[htbp]
\begin{center}
\includegraphics[width=0.98\linewidth]{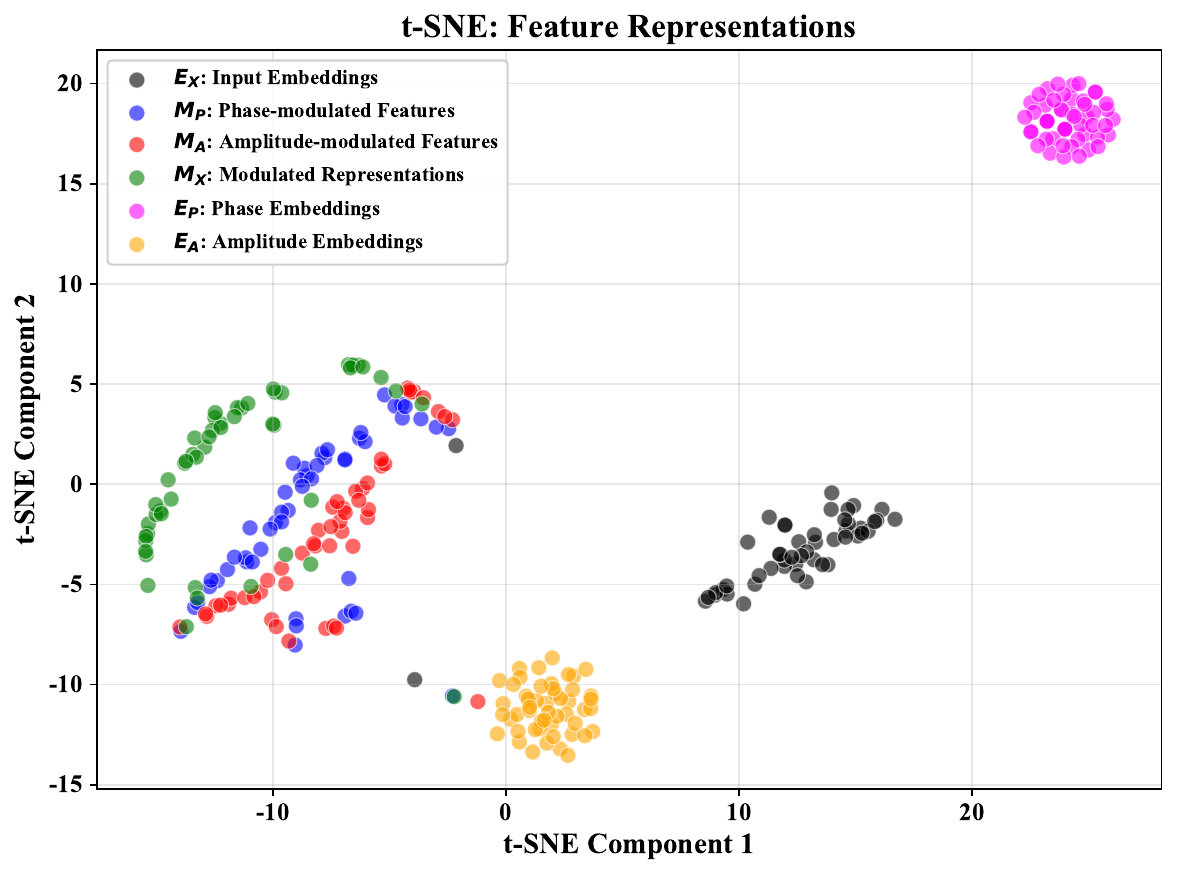}
\vspace{-0.4cm}
\caption{Visualization of the modulated representations on ECL.}
\label{fig:7}
\vspace{-0.7cm}
\end{center}
\end{figure}

\textbf{Phase-Amplitude Weights.}
Figure \ref{fig:6} visualizes the learned weights of our phase and amplitude embedding layers. The phase weights in (a) show clear cycle-dependent patterns, with weight values systematically varying across cycle positions and embedding dimensions.
The t-SNE \cite{vandermaaten08a} in (c) reveals compact, separable clusters, indicating that distinct phase states are encoded as discriminative representations. 
Meanwhile, the amplitude weights in (b) exhibit localized, channel-specific intensity variations in OT, and their t-SNE in (d) shows well-separated clusters for different channels, confirming channel-aware adaptation.
See Appendix \ref{B:2} for more cases.

\textbf{Phase-Amplitude Modulation.}
As shown in Figure \ref{fig:7}, the learnable phase embeddings $E_P$ and amplitude embeddings $E_A$ exhibit well-separated clusters, confirming their decoupling roles in feature space.
Crucially, the modulated features $M_X$ obtained by phase-modulated $M_P$ and amplitude-modulated $M_A$ aggregate into new, cohesive clusters that are clearly separated from the input features $E_X$.
Such observations demonstrate that our modulation effectively integrates phase and amplitude characteristics, generating semantically enhanced and discriminative representations that underpin its accurate forecasting performance.

\section{Conclusion}
Considering that real-world time series inherently exhibit entangled phase and amplitude characteristics within their periodic patterns, we pioneer the explicit decoupling of these components to fundamentally enhance forecasting accuracy. 
Technically, we propose PAMNet, a simple yet powerful framework to model phase-dependent positional shifts and amplitude-driven intensity variations through learnable cyclical embeddings and lightweight modulators.
Extensive experiments demonstrate that PAMNet achieves state-of-the-art performance with notable robustness and efficiency.
Our explicit phase-amplitude modulation provides a principled and effective inductive bias for time series forecasting.

\clearpage
\section*{Impact Statement}

This paper presents work whose goal is to advance the field of Machine Learning. 
There are many potential societal consequences of our work, none of which we feel must be specifically highlighted here.

\bibliography{example_paper}
\bibliographystyle{icml2025}

\newpage
\appendix
\section{More Details of PAMNet}

\begin{algorithm}[htbp]
\caption{Pseudocode of PAMNet.}
\label{alg:1}
\begin{algorithmic}[1]
\REQUIRE 
    Lookback length $\mathbf{X}\in\mathbb{R}^{L\times N}$, cycle index $c$
\ENSURE
    The prediction horizon $\hat{\mathbf{X}}\in\mathbb{R}^{H\times N}$\\
\STATE Initialize learnable embedding matrix $\Omega_p\in\mathbb{R}^{c\times d}$ and $\Omega_a\in\mathbb{R}^{c\times Nd}$ with the Xavier normal distribution
\IF{Instance Normalization is True}
\STATE $\mu, \sigma \leftarrow \text{Mean}(X), \text{STD}(X)$
\STATE $X\leftarrow \frac{X-\mu}{\sqrt{\sigma^2 + \epsilon}}$
\ENDIF
\STATE \CommentLeft{Variate Tokenization.}
\STATE $E_X=X^{\top}W+b$ \COMMENT{$E_X\in \mathbb{R}^{N\times d}$}\\
\STATE \CommentLeft{Cycle Index}
\STATE $\tau_{end}\leftarrow{\text{time stamp of the last observed step}}$
\STATE $t=\tau_{end}\mod c$
\STATE \CommentLeft{Phase Embedding}
\FOR{$i \in \{1, \cdots, N\}$}
     \STATE $E_P^i(t)\leftarrow{\text{Lookup($\Omega_p, t$)}}$
     \COMMENT{$E_P^i(t)\in\mathbb{R}^{1\times d}$}
\ENDFOR
\STATE $E_P(t)\leftarrow \text{contact}(E_p^1, \cdots, E_p^N)$
\COMMENT{$E_P(t)\in\mathbb{R}^{N\times d}$}
\STATE \CommentLeft{Amplitude Embedding}
\STATE $E_A(t)\leftarrow \text{Reshape(Lookup}(\Omega_a, t)$
\COMMENT{$E_A(t)\in \mathbb{R}^{N\times d}$}
\STATE \CommentLeft{Phase and Amplitude Modulation}
\STATE $M_P=W_3(\text{Dropout(SiLU(}W_1E_X))\odot W_2E_P(t)$
\STATE $M_A=W_3(\text{Dropout(SiLU(}W_1E_X))\odot W_2E_A(t)$\\
\STATE \CommentLeft{Forecasting Head}
\STATE $\hat{Y}=\text{MLP}(M_P+M_A)$
\STATE $\hat{\mathbf{Y}} = \hat{\mathbf{Y}}.transpose(-1, -2)$
\COMMENT{$\hat{\mathbf{Y}}\in \mathbb{R}^{H\times N}$}
\IF{Instance Normalization is True}
\STATE $\hat{Y}\leftarrow \hat{Y}\times \sqrt{\sigma^2 + \epsilon} + \mu$
\ENDIF
\end{algorithmic}
\end{algorithm}

\vspace{-0.2cm}
\subsection{Framework Implementation}
The pseudocode of PAMNet is presented in Algorithm \ref{alg:1}, where the key design lies in our learnable, decoupling phase-amplitude embedding and modulation modules. 

\vspace{-0.2cm}
\subsection{Dataset Statistics}
We evaluate the performance of PAMNet compared with various baselines on $12$ well-established benchmarks~\footnote{All the datasets are publicly available at \url{https://github.com/thuml/iTransformer}}, which are detailed in Table~\ref{tab:5}.
\begin{compactitem}
    \item \textbf{ETT} comprises two hourly-level datasets (i.e., ETTh1 and ETTh2) and two 15-minute-level datasets (i.e., ETTm1 and ETTm2). Each dataset contains seven oil and load features of electricity transformers from July 2016 to July 2018.
    \item \textbf{Electricity} encompasses the hourly electricity consumption data of 321 customers from 2012 to 2014.
    \item \textbf{Traffic} describes the road occupancy rates from the California Department of Transportation. It contains the hourly data recorded by the sensors of San Francisco freeways from 2015 to 2016.
    \item \textbf{Solar} records the solar power production of 137 PV plants in 2006, which is sampled every 10 minutes.
    \item \textbf{Weather} includes 21 indicators of weather, such as air temperature, and humidity. Its data is recorded every 10 min for 2020 in Germany.
    \item \textbf{PEMS} contains four public traffic network datasets (i.e., PEMS03, PEMS04, PEMS07, and PEMS08) in California collected by 5-minute windows.
\end{compactitem}

\vspace{-0.3cm}
\begin{table}[htbp]
\centering
\caption{Statistics information of datasets. Channel denotes the number of variates in each dataset. The dataset size of time steps is split into (Train, Validation, Test). Frequency denotes the sampling interval of time points. Period means the default cycle length.}
\label{tab:5}
\scriptsize
\begin{tabular}{c|c|c|c|c}
\specialrule{0.15em}{0pt}{1pt}
Benchmarks   & Channels & Time steps  & Frequency & Period    \\ 
\specialrule{0.10em}{1pt}{1pt}
 ETTm1   & 7  & (34465, 11521, 11521) & 15min     & 96    \\ 
\specialrule{0.10em}{1pt}{1pt}
ETTm2        & 7         & (34465, 11521, 11521) & 15min     & 96    \\ \specialrule{0.10em}{1pt}{1pt}  
 ETTh1        & 7         & (8545, 2881, 2881)    & Hourly    & 24    \\ \specialrule{0.10em}{1pt}{1pt}  
 ETTh2        & 7       & (8545, 2881, 2881)    & Hourly    & 24    \\ \specialrule{0.10em}{1pt}{1pt}
 ECL          & 321     & (18317, 2633, 5261)   & Hourly    & 168    \\ \specialrule{0.10em}{1pt}{1pt}  
 Traffic      & 862      & (12185, 1757, 3509)   & Hourly    & 168 \\ \specialrule{0.10em}{1pt}{1pt} 
Weather      & 21      & (36792, 5271, 10540)  & 10min     & 144 \\ 
\specialrule{0.10em}{1pt}{1pt}  
Solar & 137      & (36601, 5161, 10417)  & 10min     & 144    \\ 
\specialrule{0.10em}{1pt}{1pt}
 PEMS03       & 358     & (15617, 5135, 5135)   & 5min      & 288 \\ 
\specialrule{0.10em}{1pt}{1pt}
  PEMS04       & 307    & (10172, 3375, 3375)   & 5min      & 288 \\ 
  \specialrule{0.10em}{1pt}{1pt} 
 PEMS07       & 883    & (16911, 5622, 5622)   & 5min      & 288 \\ 
 \specialrule{0.10em}{1pt}{1pt} 
 PEMS08       & 170   & (10690, 3548, 3548)   & 5min      & 288 \\ 
\specialrule{0.15em}{1pt}{0pt}
\end{tabular}
\vspace{-0.2cm}
\end{table}

\subsection{Experimental Details}
PAMNet is trained for 30 epochs with early stopping based on a patience of 5 on the validation set.
During the training process, the learning rate is set to $[1\times 10^{-3}, 5\times 10^{-3}]$, the embedding dimension is set to $512$, and the dropout rate is set to 0.5 by default.
To improve training stability, we adopt a hybrid MAE loss \cite{DBLP:conf/iclr/0049PS0YY0L0T25} that operates in both the time and frequency domains:
\begin{equation}
    \begin{aligned}
        \mathcal{L}_t=\frac{1}{H}\sum_{i=1}^{H}\lvert Y&-\hat{Y}\rvert, ~
        \mathcal{L}_f=\frac{1}{H}\sum_{i=1}^{H}\lvert \mathcal{F}(Y)- \mathcal{F}(\hat{Y})\rvert, \\
        \mathcal{L} &= (1-\alpha)\times \mathcal{L}_t + \alpha\times \mathcal{L}_f,
    \end{aligned}
\end{equation}
where $\mathcal{F}$ denotes the Fast Fourier Transform, and the hyperparameter $\alpha$ is set to $[0.05, 0.35]$ for different datasets.
We use Mean Squared Error (MSE) and Mean Absolute Error (MAE) as evaluation metrics. 
Given the ground truth values $Y\in \mathbb{R}^{H\times N}$ and the predicted values $\hat{Y}\in \mathbb{R}^{H\times N}$, these metrics are defined as:
\begin{equation}
    \begin{aligned}
        \text{MSE}&=\frac{1}{H}\sum_{i=1}^{H}( Y-\hat{Y})^2, \\
        \text{MAE}&=\frac{1}{H}\sum_{i=1}^{H}\lvert Y- \hat{Y}\rvert.
    \end{aligned}
\end{equation}

\begin{table*}[htbp]
\centering
\caption{Robustness of TwinsFormer performance obtained from 5 random seeds on 12 benchmarks.}
\label{tab:6}
\scriptsize
\setlength{\tabcolsep}{1.6mm}{
\begin{tabular}{c|cc|cc|cc|cc}
\specialrule{0.15em}{0pt}{2pt}
Dataset & \multicolumn{2}{c|}{ETTm1}          & \multicolumn{2}{c|}{ETTm2}          & \multicolumn{2}{c|}{ETTh1}          & \multicolumn{2}{c}{ETTh2}             \\ 
\cmidrule(lr){1-1} \cmidrule(lr){2-3} \cmidrule(lr){4-5} \cmidrule(lr){6-7} \cmidrule(lr){8-9}
Metrics & MSE              & MAE              & MSE              & MAE              & MSE              & MAE              & MSE              & MAE              \\ 
\specialrule{0.10em}{2pt}{2pt}
96      & $0.308\pm 0.001$ & $0.347\pm 0.001$ & $0.165\pm 0.000$ & $0.243\pm 0.001$ & $0.357\pm 0.001$ & $0.383\pm 0.000$ & $0.285\pm 0.000$ & $0.330\pm 0.000$ \\
192     & $0.353\pm 0.001$ & $0.372\pm 0.001$ & $0.227\pm 0.001$ & $0.286\pm 0.000$ & $0.413\pm 0.002$ & $0.415\pm 0.001$ & $0.367\pm 0.001$ & $0.384\pm 0.001$ \\
336     & $0.375\pm 0.001$ & $0.391\pm 0.002$ & $0.285\pm 0.001$ & $0.323\pm 0.001$ & $0.453\pm 0.002$ & $0.435\pm 0.002$ & $0.414\pm 0.002$ & $0.420\pm 0.001$ \\
720     & $0.425\pm 0.002$ & $0.424\pm 0.001$ & $0.380\pm 0.002$ & $0.381\pm 0.001$ & $0.458\pm 0.003$ & $0.462\pm 0.001$ & $0.428\pm 0.002$ & $0.437\pm 0.002$ \\ 
\specialrule{0.10em}{2pt}{2pt}
Dataset & \multicolumn{2}{c|}{ECL}        & \multicolumn{2}{c|}{Traffic}       & \multicolumn{2}{c|}{Solar}   & \multicolumn{2}{c}{Weather}         \\ 
\cmidrule(lr){1-1} \cmidrule(lr){2-3} \cmidrule(lr){4-5} \cmidrule(lr){6-7} \cmidrule(lr){8-9}
Metrics & MSE              & MAE              & MSE              & MAE              & MSE              & MAE              & MSE              & MAE              \\ 
\specialrule{0.10em}{2pt}{2pt}
96      & $0.132\pm 0.001$ & $0.222\pm 0.000$ & $0.415\pm 0.001$ & $0.246\pm 0.001$ & $0.189\pm 0.002$ & $0.214\pm 0.002$ & $0.154\pm 0.000$ & $0.193\pm 0.000$ \\
192     & $0.148\pm 0.002$ & $0.238\pm 0.001$ & $0.428\pm 0.002$ & $0.258\pm 0.001$ & $0.206\pm 0.001$ & $0.226\pm 0.001$ & $0.202\pm 0.001$ & $0.239\pm 0.001$ \\
336     & $0.165\pm 0.002$ & $0.256\pm 0.001$ & $0.442\pm 0.002$ & $0.265\pm 0.001$ & $0.211\pm 0.000$ & $0.234\pm 0.001$ & $0.260\pm 0.002$ & $0.283\pm 0.001$ \\
720     & $0.199\pm 0.001$ & $0.287\pm 0.002$ & $0.475\pm 0.002$ & $0.284\pm 0.003$ & $0.211\pm 0.001$ & $0.236\pm 0.002$ & $0.343\pm 0.002$ & $0.338\pm 0.002$ \\ 
\specialrule{0.10em}{2pt}{2pt}
Dataset & \multicolumn{2}{c|}{PEMS03}         & \multicolumn{2}{c|}{PEMS04}         & \multicolumn{2}{c|}{PEMS07}         & \multicolumn{2}{c}{PEMS08}          \\ 
\cmidrule(lr){1-1} \cmidrule(lr){2-3} \cmidrule(lr){4-5} \cmidrule(lr){6-7} \cmidrule(lr){8-9}
Metrics & MSE              & MAE              & MSE              & MAE              & MSE              & MAE              & MSE              & MAE              \\ 
\specialrule{0.10em}{2pt}{2pt}
12      & $0.059\pm 0.000$ & $0.157\pm 0.000$ & $0.067\pm 0.001$ & $0.163\pm 0.001$ & $0.055\pm 0.000$ & $0.142\pm 0.000$ & $0.072\pm 0.000$ & $0.167\pm 0.001$ \\
24      & $0.072\pm 0.000$ & $0.173\pm 0.000$ & $0.075\pm 0.001$ & $0.171\pm 0.000$ & $0.065\pm 0.000$ & $0.153\pm 0.000$ & $0.095\pm 0.000$ & $0.188\pm 0.000$ \\
48      & $0.098\pm 0.001$ & $0.197\pm 0.001$ & $0.090\pm 0.002$ & $0.186\pm 0.001$ & $0.084\pm 0.001$ & $0.169\pm 0.001$ & $0.140\pm 0.001$ & $0.222\pm 0.001$ \\
96      & $0.134\pm 0.002$ & $0.225\pm 0.001$ & $0.106\pm 0.001$ & $0.201\pm 0.001$ & $0.109\pm 0.001$ & $0.187\pm 0.001$ & $0.239\pm 0.002$ & $0.267\pm 0.001$ \\ 
\specialrule{0.15em}{2pt}{0pt}
\end{tabular}}
\end{table*}

\section{More Results of PAMNet}
\subsection{Error Bars}
We calculate the standard deviation of PAMNet performance by training the model with five different random seeds across 12 datasets.
As seen in Table~\ref{tab:6}, the error bars of all the results are tiny, indicating our PAMNet is robust and reliable.

\subsection{More Visual Cases}
\label{B:2}

\begin{figure}[htbp]
\begin{center}
\includegraphics[width=1.0\linewidth]{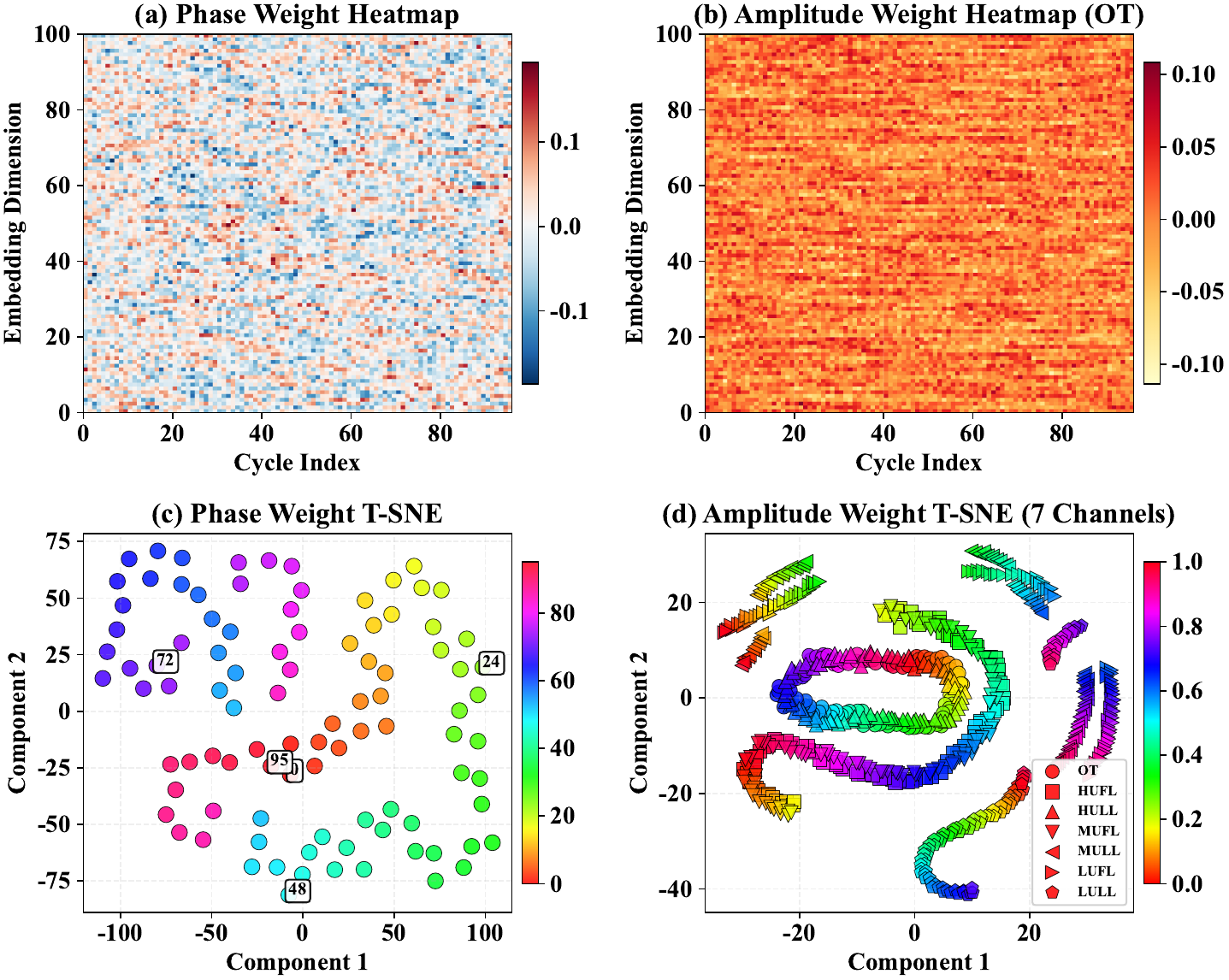}
\vspace{-0.8cm}
\caption{The learned phase and amplitude weights on ETTm1.}
\label{fig:8}
\vspace{-0.4cm}
\end{center}
\end{figure}

\begin{figure}[!ht]
\begin{center}
\includegraphics[width=1.0\linewidth]{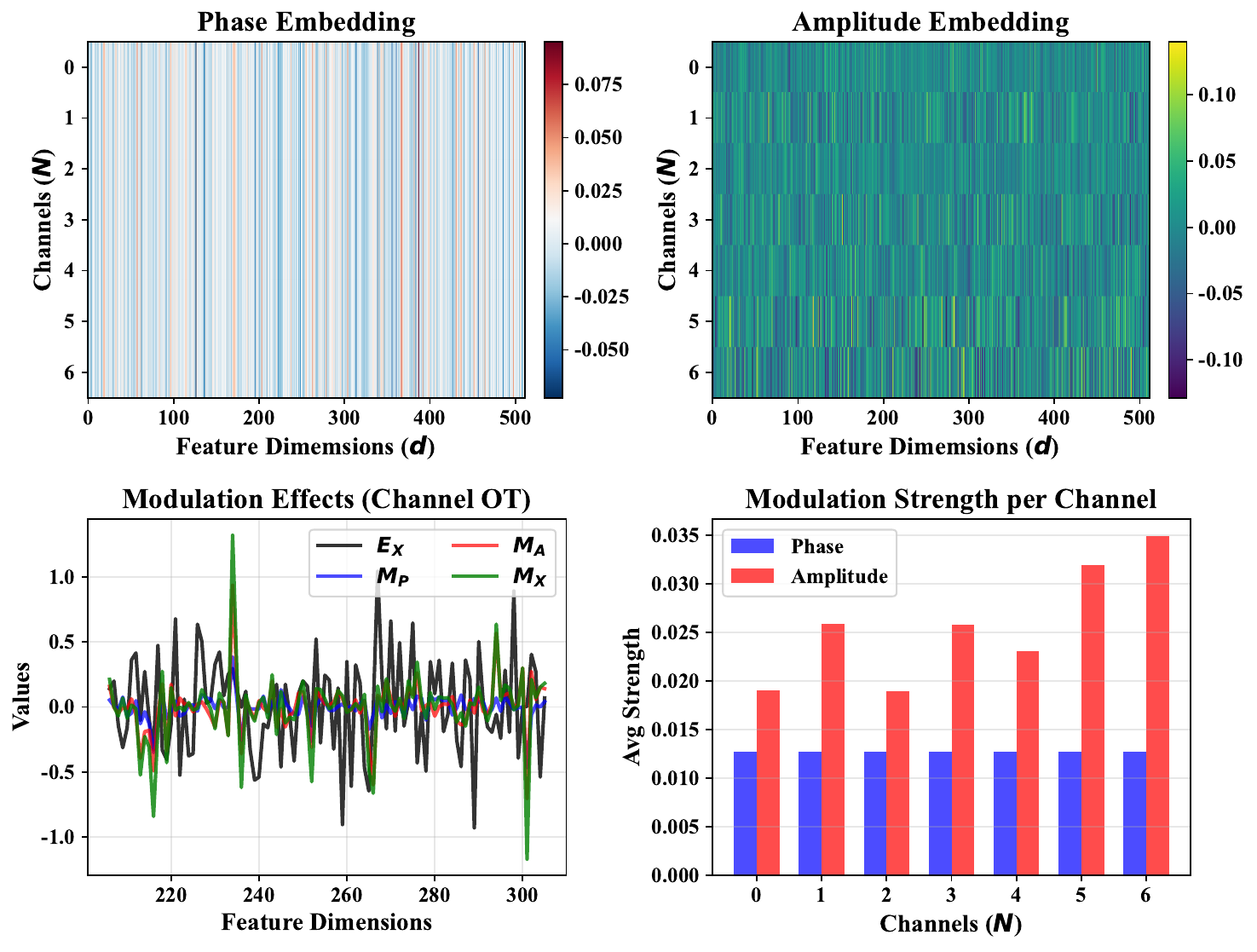}
\vspace{-0.8cm}
\caption{The modulated representations on ETTm1.}
\label{fig:9}
\vspace{-0.4cm}
\end{center}
\end{figure}

\begin{figure}[!ht]
\begin{center}
\includegraphics[width=0.90\linewidth]{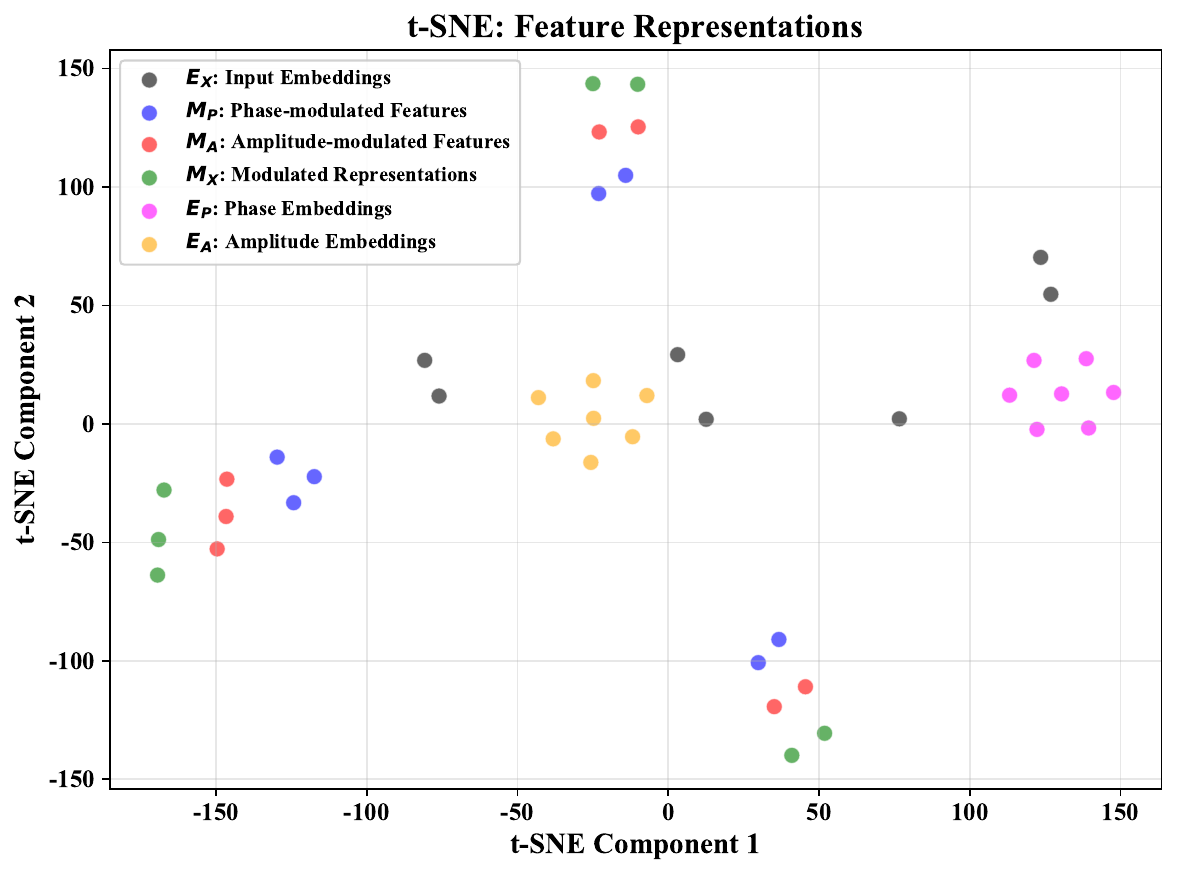}
\vspace{-0.3cm}
\caption{t-SNE visualization of the learned features on ETTm1.}
\label{fig:10}
\vspace{-0.4cm}
\end{center}
\end{figure}

As shown in Figures \ref{fig:8}, \ref{fig:9}, and \ref{fig:10}, these visualizations collectively validate PAMNet’s core design: cycle‑aware embeddings encode periodic dynamics, dual‑path modulation enhances feature discriminability, and complementary fusion yields semantically rich temporal representations. 
By bridging signal processing principles with deep learning, PAMNet provides an interpretable and efficient framework for modeling complex time series patterns.

\subsection{Full Comparison Results}
Table \ref{tab:7} provides the full comparison results of PAMNet against $9$ baselines across $12$ real-world datasets.

\begin{table*}[htbp]
\centering
\vspace{-0.4cm}
\caption{Full results of multivariate time series forecasting across 12 benchmarks. The best results are highlighted in red bold, the second best are underlined in blue, and the \textbf{Count} row counts the number of times each model ranks in the top 1.}
\label{tab:7}
\vspace{0.1cm}
\scriptsize
\setlength{\tabcolsep}{1.0mm}
{
\begin{tabular}{cc|cc|cc|cc|cc|cc|cc|cc|cc|cc|cc}
\specialrule{0.15em}{0pt}{2pt}
\multicolumn{2}{c|}{\multirow{2}{*}{Model}}         & \multicolumn{2}{c|}{PAMNet}     & \multicolumn{2}{c|}{TQNet}   & \multicolumn{2}{c|}{FilterTS} & \multicolumn{2}{c|}{Amplifier}  & \multicolumn{2}{c|}{CycleNet} & \multicolumn{2}{c|}{TimeMixer}  & \multicolumn{2}{c|}{iTransformer} & \multicolumn{2}{c|}{PatchTST} & \multicolumn{2}{c|}{TimesNet} & \multicolumn{2}{c}{Dlinear} \\ 
\cmidrule(lr){3-4} \cmidrule(lr){5-6} \cmidrule(lr){7-8} \cmidrule(lr){9-10} \cmidrule(lr){11-12} \cmidrule(lr){13-14} \cmidrule(lr){15-16} \cmidrule(lr){17-18} \cmidrule(lr){19-20} \cmidrule(lr){21-22}
\multicolumn{2}{c|}{}                               & \multicolumn{2}{c|}{(Ours)}     & \multicolumn{2}{c|}{(\citeyear{lin2025TQNet})}    & \multicolumn{2}{c|}{(\citeyear{DBLP:conf/aaai/WangLDW25})}     & \multicolumn{2}{c|}{(\citeyear{DBLP:conf/aaai/Fei000N25})}      & \multicolumn{2}{c|}{(\citeyear{DBLP:conf/nips/Lin0HWMZ24})}     & \multicolumn{2}{c|}{(\citeyear{DBLP:conf/iclr/WangWSHLMZ024})}      & \multicolumn{2}{c|}{(\citeyear{DBLP:conf/iclr/LiuHZWWML24})}         & \multicolumn{2}{c|}{(\citeyear{DBLP:conf/iclr/NieNSK23})}     & \multicolumn{2}{c|}{(\citeyear{DBLP:conf/iclr/WuHLZ0L23})}     & \multicolumn{2}{c}{(\citeyear{DBLP:conf/aaai/ZengCZ023})}    \\  
\specialrule{0.10em}{1pt}{1pt}
\multicolumn{2}{c|}{Metric}                         & MSE            & MAE            & MSE            & MAE         & MSE           & MAE           & MSE            & MAE            & MSE             & MAE         & MSE            & MAE            & MSE             & MAE             & MSE           & MAE           & MSE           & MAE           & MSE          & MAE          \\ 
\specialrule{0.10em}{1pt}{1pt}
\multicolumn{1}{c|}{\multirow{5}{*}{\rotatebox{90}{ETTm1}}}   & 96  & {\color{red}\textbf{0.308}} & {\color{red}\textbf{0.347}} & {\color{blue}\underline{0.311}}    & {\color{blue}\underline{0.353}} & 0.321         & 0.360         & 0.316          & 0.355          & 0.319           & 0.360       & 0.320          & 0.357          & 0.334           & 0.368           & 0.329         & 0.367         & 0.338         & 0.375         & 0.345        & 0.372        \\
\multicolumn{1}{c|}{}                         & 192 & {\color{red}\textbf{0.353}} & {\color{red}\textbf{0.372}} & {\color{blue}\underline{0.356}}    & {\color{blue}\underline{0.378}} & 0.363         & 0.382         & 0.361          & 0.381          & 0.360           & 0.381       & 0.361          & 0.381          & 0.377           & 0.391           & 0.367         & 0.385         & 0.374         & 0.387         & 0.380        & 0.389        \\
\multicolumn{1}{c|}{}                         & 336 & {\color{red}\textbf{0.375}} & {\color{red}\textbf{0.391}} & 0.390          & {\color{blue}\underline{0.401}} & 0.396         & 0.404         & 0.393          & 0.404          & 0.389           & 0.403       & 0.390          & 0.404          & 0.426           & 0.420           & 0.399         & 0.410         & 0.410         & 0.411         & 0.413        & 0.413        \\
\multicolumn{1}{c|}{}                         & 720 & {\color{red}\textbf{0.425}} & {\color{red}\textbf{0.424}} & {\color{blue}\underline{0.452}}    & 0.440       & 0.462         & {\color{blue}\underline{0.438}}   & 0.456          & 0.440          & 0.447           & 0.441       & 0.454          & 0.441          & 0.491           & 0.459           & 0.454         & 0.439         & 0.478         & 0.450         & 0.474        & 0.453        \\ 
\cmidrule(lr){2-22}
\multicolumn{1}{c|}{}                         & Avg & {\color{red}\textbf{0.365}} & {\color{red}\textbf{0.384}} & {\color{blue}\underline{0.377}}    & {\color{blue}\underline{0.393}} & 0.385         & 0.396         & 0.382          & 0.395          & 0.379           & 0.396       & 0.381          & 0.395          & 0.407           & 0.410           & 0.387         & 0.400         & 0.400         & 0.406         & 0.403        & 0.407        \\ 
\specialrule{0.10em}{1pt}{1pt}
\multicolumn{1}{c|}{\multirow{5}{*}{\rotatebox{90}{ETTm2}}}   & 96  & {\color{blue}\underline{0.165}}    & {\color{red}\textbf{0.243}} & 0.173          & 0.256       & 0.173         & 0.256         & 0.178          & 0.261          & {\color{red}\textbf{0.163}}  & {\color{blue}\underline{0.246}} & 0.175          & 0.258          & 0.180           & 0.264           & 0.175         & 0.259         & 0.187         & 0.267         & 0.193        & 0.292        \\
\multicolumn{1}{c|}{}                         & 192 & {\color{red}\textbf{0.227}} & {\color{red}\textbf{0.286}} & 0.238          & 0.298       & 0.238         & 0.299         & 0.244          & 0.304          & {\color{blue}\underline{0.229}}     & {\color{blue}\underline{0.290}} & 0.237          & 0.299          & 0.250           & 0.309           & 0.241         & 0.302         & 0.249         & 0.309         & 0.284        & 0.362        \\
\multicolumn{1}{c|}{}                         & 336 & {\color{blue}\underline{0.285}}    & {\color{red}\textbf{0.323}} & 0.301          & 0.340       & 0.300         & 0.338         & 0.309          & 0.346          & {\color{red}\textbf{0.284}}  & {\color{blue}\underline{0.327}} & 0.298          & 0.340          & 0.311           & 0.348           & 0.305         & 0.343         & 0.321         & 0.351         & 0.369        & 0.427        \\
\multicolumn{1}{c|}{}                         & 720 & {\color{red}\textbf{0.380}} & {\color{red}\textbf{0.381}} & 0.397          & 0.396       & 0.399         & 0.395         & 0.390          & 0.394          & {\color{blue}\underline{0.389}}     & {\color{blue}\underline{0.391}} & 0.391          & 0.396          & 0.412           & 0.407           & 0.402         & 0.400         & 0.408         & 0.403         & 0.554        & 0.522        \\ 
\cmidrule(lr){2-22}
\multicolumn{1}{c|}{}                         & Avg & {\color{red}\textbf{0.264}} & {\color{red}\textbf{0.308}} & 0.277          & 0.323       & 0.277         & 0.322         & 0.280          & 0.326          & {\color{blue}\underline{0.266}}     & {\color{blue}\underline{0.314}} & 0.275          & 0.323          & 0.288           & 0.332           & 0.281         & 0.326         & 0.291         & 0.333         & 0.350        & 0.401        \\ 
\specialrule{0.10em}{1pt}{1pt}
\multicolumn{1}{c|}{\multirow{5}{*}{\rotatebox{90}{ETTh1}}}   & 96  & {\color{red}\textbf{0.357}} & {\color{red}\textbf{0.383}} & {\color{blue}\underline{0.371}}          & 0.393 & 0.375         & {\color{blue}\underline{0.391}}         & {\color{blue}\underline{0.371}}          & 0.392          & 0.375           & 0.395       & 0.375          & 0.400          & 0.386           & 0.405           & 0.414         & 0.419         & 0.384         & 0.402         & 0.386        & 0.400        \\
\multicolumn{1}{c|}{}                         & 192 & {\color{red}\textbf{0.413}} & {\color{red}\textbf{0.415}} & 0.428          & 0.426       & {\color{blue}\underline{0.424}}   & {\color{blue}\underline{0.421}}   & 0.425          & 0.422          & 0.436           & 0.428       & 0.429          & 0.421          & 0.441           & 0.436           & 0.460         & 0.445         & 0.436         & 0.429         & 0.437        & 0.432        \\
\multicolumn{1}{c|}{}                         & 336 & {\color{blue}\underline{0.453}}    & {\color{blue}\underline{0.435}}    & 0.476          & 0.446       & 0.465         & 0.442         & {\color{red}\textbf{0.448}} & {\color{red}\textbf{0.434}} & 0.496           & 0.455       & 0.484          & 0.458          & 0.487           & 0.458           & 0.501         & 0.466         & 0.491         & 0.469         & 0.481        & 0.459        \\
\multicolumn{1}{c|}{}                         & 720 & {\color{red}\textbf{0.458}} & {\color{red}\textbf{0.462}} & 0.487          & 0.470       & {\color{blue}\underline{0.472}}   & 0.466         & 0.476          & {\color{blue}\underline{0.464}}    & 0.520           & 0.484       & 0.498          & 0.482          & 0.503           & 0.491           & 0.500         & 0.488         & 0.521         & 0.500         & 0.519        & 0.516        \\ 
\cmidrule(lr){2-22}
\multicolumn{1}{c|}{}                         & Avg & {\color{red}\textbf{0.420}} & {\color{red}\textbf{0.424}} & 0.441          & 0.434       & 0.434         & 0.430         & {\color{blue}\underline{0.430}}    & {\color{blue}\underline{0.428}}    & 0.457           & 0.441       & 0.447          & 0.440          & 0.454           & 0.448           & 0.469         & 0.455         & 0.458         & 0.450         & 0.456        & 0.452        \\ 
\specialrule{0.10em}{1pt}{1pt}
\multicolumn{1}{c|}{\multirow{5}{*}{\rotatebox{90}{ETTh2}}}   & 96  & {\color{red}\textbf{0.285}} & {\color{red}\textbf{0.330}} & 0.295          & 0.343       & 0.290         & {\color{blue}\underline{0.338}}   & 0.290          & 0.341          & 0.298           & 0.344       & {\color{blue}\underline{0.289}}    & 0.341          & 0.297           & 0.349           & 0.302         & 0.348         & 0.340         & 0.374         & 0.333        & 0.387        \\
\multicolumn{1}{c|}{}                         & 192 & {\color{red}\textbf{0.367}} & {\color{red}\textbf{0.384}} & {\color{red}\textbf{0.367}} & 0.393       & 0.374         & {\color{blue}\underline{0.390}}   & {\color{blue}\underline{0.369}}    & {\color{blue}\underline{0.390}}    & 0.372           & 0.396       & 0.372          & 0.392          & 0.380           & 0.400           & 0.388         & 0.400         & 0.402         & 0.414         & 0.477        & 0.476        \\
\multicolumn{1}{c|}{}                         & 336 & {\color{blue}\underline{0.414}}    & {\color{blue}\underline{0.420}}    & 0.417          & 0.427       & 0.415         & 0.424         & 0.419          & 0.431          & 0.431           & 0.439       & {\color{red}\textbf{0.386}} & {\color{red}\textbf{0.414}} & 0.428           & 0.432           & 0.426         & 0.433         & 0.452         & 0.452         & 0.594        & 0.541        \\
\multicolumn{1}{c|}{}                         & 720 & 0.428          & {\color{blue}\underline{0.437}}    & 0.433          & 0.446       & {\color{blue}\underline{0.420}}   & 0.438         & 0.446          & 0.456          & 0.450           & 0.458       & {\color{red}\textbf{0.412}} & {\color{red}\textbf{0.434}} & {\color{blue}\underline{0.427}}     & 0.445           & 0.431         & 0.446         & 0.462         & 0.468         & 0.831        & 0.657        \\ 
\cmidrule(lr){2-22}
\multicolumn{1}{c|}{}                         & Avg & {\color{blue}\underline{0.374}}    & {\color{red}\textbf{0.393}} & 0.378          & 0.402       & 0.375         & 0.398         & 0.381          & 0.405          & 0.388           & 0.409       & {\color{red}\textbf{0.364}} & {\color{blue}\underline{0.395}}    & 0.383           & 0.407           & 0.387         & 0.407         & 0.414         & 0.427         & 0.559        & 0.515        \\ 
\specialrule{0.10em}{1pt}{1pt}
\multicolumn{1}{c|}{\multirow{5}{*}{\rotatebox{90}{ECL}}}     & 96  & {\color{red}\textbf{0.132}} & {\color{red}\textbf{0.222}} & {\color{blue}\underline{0.134}}    & {\color{blue}\underline{0.229}} & 0.151         & 0.245         & 0.147          & 0.242          & 0.136           & {\color{blue}\underline{0.229}} & 0.153          & 0.247          & 0.148           & 0.240           & 0.181         & 0.270         & 0.168         & 0.272         & 0.197        & 0.282        \\
\multicolumn{1}{c|}{}                         & 192 & {\color{red}\textbf{0.148}} & {\color{red}\textbf{0.238}} & 0.154          & 0.247       & 0.164         & 0.256         & 0.158          & 0.251          & {\color{blue}\underline{0.152}}     & {\color{blue}\underline{0.244}} & 0.166          & 0.256          & 0.162           & 0.253           & 0.188         & 0.274         & 0.184         & 0.289         & 0.196        & 0.285        \\
\multicolumn{1}{c|}{}                         & 336 & {\color{red}\textbf{0.165}} & {\color{red}\textbf{0.256}} & {\color{blue}\underline{0.169}}    & {\color{blue}\underline{0.264}} & 0.181         & 0.274         & 0.175          & 0.271          & 0.170           & {\color{blue}\underline{0.264}} & 0.185          & 0.277          & 0.178           & 0.269           & 0.204         & 0.293         & 0.198         & 0.300         & 0.209        & 0.301        \\
\multicolumn{1}{c|}{}                         & 720 & {\color{red}\textbf{0.199}} & {\color{red}\textbf{0.287}} & {\color{blue}\underline{0.201}}    & {\color{blue}\underline{0.294}} & 0.225         & 0.311         & 0.206          & 0.298          & 0.212           & 0.299       & 0.225          & 0.310          & 0.225           & 0.317           & 0.246         & 0.324         & 0.220         & 0.320         & 0.245        & 0.333        \\ 
\cmidrule(lr){2-22}
\multicolumn{1}{c|}{}                         & Avg & {\color{red}\textbf{0.161}} & {\color{red}\textbf{0.251}} & {\color{blue}\underline{0.164}}    & {\color{blue}\underline{0.259}}       & 0.180         & 0.272         & 0.172          & 0.266          & 0.168           & {\color{blue}\underline{0.259}}       & 0.182          & 0.272          & 0.178           & 0.270           & 0.205         & 0.290         & 0.193         & 0.295         & 0.212        & 0.300        \\ 
\specialrule{0.10em}{1pt}{1pt}
\multicolumn{1}{c|}{\multirow{5}{*}{\rotatebox{90}{Traffic}}} & 96  & 0.415          & {\color{red}\textbf{0.246}} & {\color{blue}\underline{0.413}}    & {\color{blue}\underline{0.261}} & 0.446         & 0.308         & 0.456          & 0.299          & 0.458           & 0.296       & 0.462          & 0.285          & {\color{red}\textbf{0.395}}  & 0.268           & 0.462         & 0.290         & 0.593         & 0.321         & 0.650        & 0.396        \\
\multicolumn{1}{c|}{}                         & 192 & {\color{blue}\underline{0.428}}    & {\color{red}\textbf{0.258}} & 0.432          & {\color{blue}\underline{0.271}} & 0.456         & 0.308         & 0.472          & 0.318          & 0.457           & 0.294       & 0.473          & 0.296          & {\color{red}\textbf{0.417}}  & 0.276           & 0.466         & 0.290         & 0.617         & 0.336         & 0.598        & 0.370        \\
\multicolumn{1}{c|}{}                         & 336 & {\color{blue}\underline{0.442}}    & {\color{red}\textbf{0.265}} & 0.450          & {\color{blue}\underline{0.277}} & 0.472         & 0.313         & 0.487          & 0.320          & 0.470           & 0.299       & 0.498          & 0.296          & {\color{red}\textbf{0.433}}  & 0.283           & 0.482         & 0.300         & 0.629         & 0.336         & 0.605        & 0.373        \\
\multicolumn{1}{c|}{}                         & 720 & {\color{blue}\underline{0.475}}    & {\color{red}\textbf{0.284}} & 0.486          & {\color{blue}\underline{0.295}} & 0.508         & 0.332         & 0.517          & 0.332          & 0.502           & 0.314       & 0.506          & 0.313          & {\color{red}\textbf{0.467}}  & 0.302           & 0.514         & 0.320         & 0.640         & 0.350         & 0.645        & 0.394        \\ 
\cmidrule(lr){2-22}
\multicolumn{1}{c|}{}                         & Avg & {\color{blue}\underline{0.440}}    & {\color{red}\textbf{0.263}} & 0.445          & {\color{blue}\underline{0.276}} & 0.470         & 0.315         & 0.483          & 0.317          & 0.472           & 0.314       & 0.484          & 0.297          & {\color{red}\textbf{0.428}}  & 0.282           & 0.481         & 0.300         & 0.620         & 0.336         & 0.625        & 0.383        \\ 
\specialrule{0.10em}{1pt}{1pt}
\multicolumn{1}{c|}{\multirow{5}{*}{\rotatebox{90}{Weather}}} & 96  & {\color{red}\textbf{0.154}} & {\color{red}\textbf{0.193}} & {\color{blue}\underline{0.157}}    & {\color{blue}\underline{0.200}} & 0.162         & 0.207         & 0.167          & 0.212          & 0.158           & 0.203       & 0.163          & 0.209          & 0.174           & 0.214           & 0.177         & 0.210         & 0.172         & 0.220         & 0.196        & 0.255        \\
\multicolumn{1}{c|}{}                         & 192 & {\color{red}\textbf{0.202}} & {\color{red}\textbf{0.239}} & {\color{blue}\underline{0.206}}    & {\color{blue}\underline{0.245}} & 0.209         & 0.252         & 0.215          & 0.251          & 0.207           & 0.247       & 0.208          & 0.250          & 0.221           & 0.254           & 0.225         & 0.250         & 0.219         & 0.261         & 0.237        & 0.296        \\
\multicolumn{1}{c|}{}                         & 336 & {\color{red}\textbf{0.260}} & {\color{red}\textbf{0.283}} & {\color{blue}\underline{0.262}}    & {\color{blue}\underline{0.287}} & 0.263         & 0.292         & 0.276          & 0.292          & 0.262           & 0.289       & 0.251          & 0.287          & 0.278           & 0.296           & 0.278         & 0.290         & 0.280         & 0.306         & 0.283        & 0.335        \\
\multicolumn{1}{c|}{}                         & 720 & {\color{red}\textbf{0.343}} & {\color{red}\textbf{0.338}} & {\color{blue}\underline{0.344}}    & {\color{blue}\underline{0.342}} & 0.345         & 0.344         & 0.352          & 0.346          & 0.344           & 0.344       & 0.339          & 0.341          & 0.358           & 0.349           & 0.354         & 0.340         & 0.365         & 0.359         & 0.345        & 0.381        \\ 
\cmidrule(lr){2-22}
\multicolumn{1}{c|}{}                         & Avg & {\color{red}\textbf{0.240}} & {\color{red}\textbf{0.263}} & {\color{blue}\underline{0.242}}    & {\color{blue}\underline{0.269}} & 0.245         & 0.274         & 0.253          & 0.275          & 0.243           & 0.271       & 0.240          & 0.271          & 0.258           & 0.278           & 0.259         & 0.273         & 0.259         & 0.287         & 0.265        & 0.317        \\ 
\specialrule{0.10em}{1pt}{1pt}
\multicolumn{1}{c|}{\multirow{5}{*}{\rotatebox{90}{Solar}}}   & 96  & {\color{blue}\underline{0.189}}    & {\color{red}\textbf{0.214}} & {\color{red}\textbf{0.173}} & {\color{blue}\underline{0.233}} & 0.196         & 0.264         & 0.234          & 0.283          & 0.190           & 0.247       & {\color{blue}\underline{0.189}}    & 0.259          & 0.203           & 0.237           & 0.234         & 0.286         & 0.250         & 0.292         & 0.290        & 0.378        \\
\multicolumn{1}{c|}{}                         & 192 & {\color{blue}\underline{0.206}}    & {\color{red}\textbf{0.226}} & {\color{red}\textbf{0.199}} & {\color{blue}\underline{0.257}} & 0.211         & 0.278         & 0.237          & 0.259          & 0.210           & 0.266       & 0.222          & 0.283          & 0.233           & 0.261           & 0.267         & 0.310         & 0.296         & 0.318         & 0.320        & 0.398        \\
\multicolumn{1}{c|}{}                         & 336 & {\color{red}\textbf{0.211}} & {\color{red}\textbf{0.234}} & {\color{red}\textbf{0.211}} & {\color{blue}\underline{0.263}} & 0.226         & 0.284         & 0.247          & 0.269          & 0.217           & 0.266       & 0.231          & 0.292          & 0.248           & 0.273           & 0.290         & 0.315         & 0.319         & 0.330         & 0.353        & 0.415        \\
\multicolumn{1}{c|}{}                         & 720 & {\color{blue}\underline{0.211}}    & {\color{red}\textbf{0.236}} & {\color{red}\textbf{0.209}} & 0.270       & 0.227         & 0.282         & 0.246          & 0.270          & 0.223           & {\color{blue}\underline{0.266}} & 0.223          & 0.285          & 0.249           & 0.275           & 0.289         & 0.317         & 0.338         & 0.337         & 0.356        & 0.413        \\ 
\cmidrule(lr){2-22}
\multicolumn{1}{c|}{}                         & Avg & {\color{blue}\underline{0.204}}    & {\color{red}\textbf{0.228}} & {\color{red}\textbf{0.198}} & {\color{blue}\underline{0.256}} & 0.215         & 0.277         & 0.241          & 0.270          & 0.210           & 0.261       & 0.216          & 0.280          & 0.233           & 0.262           & 0.270         & 0.307         & 0.301         & 0.319         & 0.330        & 0.401        \\ 
\specialrule{0.10em}{1pt}{1pt}
\multicolumn{1}{c|}{\multirow{5}{*}{\rotatebox{90}{PEMS03}}}  & 12  & {\color{red}\textbf{0.059}} & {\color{red}\textbf{0.157}} & {\color{blue}\underline{0.060}}    & {\color{blue}\underline{0.161}} & 0.072         & 0.181         & 0.070          & 0.172          & 0.066           & 0.172       & 0.076          & 0.188          & 0.071           & 0.174           & 0.099         & 0.216         & 0.085         & 0.192         & 0.122        & 0.243        \\
\multicolumn{1}{c|}{}                         & 24  & {\color{red}\textbf{0.072}} & {\color{red}\textbf{0.173}} & {\color{blue}\underline{0.077}}    & {\color{blue}\underline{0.182}} & 0.104         & 0.219         & 0.090          & 0.200          & 0.089           & 0.201       & 0.113          & 0.226          & 0.093           & 0.201           & 0.142         & 0.259         & 0.118         & 0.223         & 0.201        & 0.317        \\
\multicolumn{1}{c|}{}                         & 48  & {\color{red}\textbf{0.098}} & {\color{red}\textbf{0.197}} & {\color{blue}\underline{0.104}}    & {\color{blue}\underline{0.215}} & 0.155         & 0.269         & 0.147          & 0.260          & 0.136           & 0.247       & 0.191          & 0.292          & 0.125           & 0.236           & 0.211         & 0.319         & 0.155         & 0.260         & 0.333        & 0.425        \\
\multicolumn{1}{c|}{}                         & 96  & {\color{red}\textbf{0.134}} & {\color{red}\textbf{0.225}} & {\color{blue}\underline{0.148}}    & {\color{blue}\underline{0.253}} & 0.203         & 0.315         & 0.217          & 0.323          & 0.182           & 0.282       & 0.288          & 0.363          & 0.164           & 0.275           & 0.269         & 0.370         & 0.228         & 0.317         & 0.457        & 0.515        \\ 
\cmidrule(lr){2-22}
\multicolumn{1}{c|}{}                         & Avg & {\color{red}\textbf{0.091}} & {\color{red}\textbf{0.188}} & {\color{blue}\underline{0.097}}    & {\color{blue}\underline{0.203}} & 0.134         & 0.246         & 0.131          & 0.239          & 0.118           & 0.226       & 0.167          & 0.267          & 0.113           & 0.222           & 0.180         & 0.291         & 0.147         & 0.248         & 0.278        & 0.375        \\ 
\specialrule{0.10em}{1pt}{1pt}
\multicolumn{1}{c|}{\multirow{5}{*}{\rotatebox{90}{PEMS04}}}  & 12  & {\color{red}\textbf{0.067}} & {\color{red}\textbf{0.163}} & {\color{red}\textbf{0.067}} & {\color{blue}\underline{0.166}} & 0.087         & 0.199         & 0.082          & 0.190          & {\color{blue}\underline{0.078}}           & 0.186       & 0.092          & 0.204          & {\color{blue}\underline{0.078}}           & 0.183           & 0.105         & 0.224         & 0.087         & 0.195         & 0.148        & 0.272        \\
\multicolumn{1}{c|}{}                         & 24  & {\color{red}\textbf{0.075}} & {\color{red}\textbf{0.171}} & {\color{blue}\underline{0.077}}    & {\color{blue}\underline{0.181}} & 0.107         & 0.223         & 0.102          & 0.215          & 0.099           & 0.212       & 0.128          & 0.243          & 0.095           & 0.205           & 0.153         & 0.275         & 0.103         & 0.215         & 0.224        & 0.340        \\
\multicolumn{1}{c|}{}                         & 48  & {\color{red}\textbf{0.090}} & {\color{red}\textbf{0.186}} & {\color{blue}\underline{0.097}}    & {\color{blue}\underline{0.206}} & 0.138         & 0.255         & 0.151          & 0.269          & 0.133           & 0.248       & 0.213          & 0.315          & 0.120           & 0.233           & 0.229         & 0.339         & 0.136         & 0.250         & 0.355        & 0.437        \\
\multicolumn{1}{c|}{}                         & 96  & {\color{red}\textbf{0.106}} & {\color{red}\textbf{0.201}} & {\color{blue}\underline{0.123}}    & {\color{blue}\underline{0.233}} & 0.166         & 0.285         & 0.205          & 0.320          & 0.167           & 0.281       & 0.307          & 0.384          & 0.150           & 0.262           & 0.291         & 0.389         & 0.190         & 0.303         & 0.452        & 0.504        \\ 
\cmidrule(lr){2-22}
\multicolumn{1}{c|}{}                         & Avg & {\color{red}\textbf{0.085}} & {\color{red}\textbf{0.180}} & {\color{blue}\underline{0.091}}    & {\color{blue}\underline{0.197}} & 0.125         & 0.241         & 0.135          & 0.249          & 0.119           & 0.232       & 0.185          & 0.287          & 0.111           & 0.221           & 0.195         & 0.307         & 0.129         & 0.241         & 0.295        & 0.388        \\ 
\specialrule{0.10em}{1pt}{1pt}
\multicolumn{1}{c|}{\multirow{5}{*}{\rotatebox{90}{PEMS07}}}  & 12  & {\color{blue}\underline{0.055}}    & {\color{red}\textbf{0.142}} & {\color{red}\textbf{0.051}} & {\color{blue}\underline{0.143}} & 0.067         & 0.170         & 0.079          & 0.179          & 0.062           & 0.162       & 0.073          & 0.184          & 0.067           & 0.265           & 0.095         & 0.207         & 0.082         & 0.181         & 0.115        & 0.242        \\
\multicolumn{1}{c|}{}                         & 24  & {\color{blue}\underline{0.065}}    & {\color{red}\textbf{0.153}} & {\color{red}\textbf{0.063}} & {\color{blue}\underline{0.159}} & 0.095         & 0.203         & 0.094          & 0.196          & 0.086           & 0.192       & 0.111          & 0.219          & 0.088           & 0.190           & 0.150         & 0.262         & 0.101         & 0.204         & 0.210        & 0.329        \\
\multicolumn{1}{c|}{}                         & 48  & {\color{blue}\underline{0.084}}    & {\color{red}\textbf{0.169}} & {\color{red}\textbf{0.081}} & {\color{blue}\underline{0.179}} & 0.146         & 0.239         & 0.129          & 0.237          & 0.128           & 0.234       & 0.237          & 0.328          & 0.110           & 0.215           & 0.253         & 0.340         & 0.134         & 0.238         & 0.398        & 0.458        \\
\multicolumn{1}{c|}{}                         & 96  & {\color{blue}\underline{0.109}}    & {\color{red}\textbf{0.187}} & {\color{red}\textbf{0.103}} & {\color{blue}\underline{0.203}} & 0.172         & 0.266         & 0.185          & 0.291          & 0.176           & 0.268       & 0.303          & 0.354          & 0.139           & 0.245           & 0.346         & 0.404         & 0.181         & 0.279         & 0.594        & 0.553        \\ 
\cmidrule(lr){2-22}
\multicolumn{1}{c|}{}                         & Avg & {\color{blue}\underline{0.078}}    & {\color{red}\textbf{0.163}} & {\color{red}\textbf{0.075}} & {\color{blue}\underline{0.171}} & 0.120         & 0.220         & 0.122          & 0.226          & 0.113           & 0.214       & 0.181          & 0.271          & 0.101           & 0.204           & 0.211         & 0.303         & 0.125         & 0.226         & 0.329        & 0.396        \\ 
\specialrule{0.10em}{1pt}{1pt}
\multicolumn{1}{c|}{\multirow{5}{*}{\rotatebox{90}{PEMS08}}}  & 12  & {\color{blue}\underline{0.072}}    & {\color{red}\textbf{0.167}} & {\color{red}\textbf{0.071}} & {\color{blue}\underline{0.170}} & 0.086         & 0.194         & 0.079          & 0.182          & 0.082           & 0.185       & 0.091          & 0.201          & 0.079           & 0.182           & 0.168         & 0.232         & 0.112         & 0.212         & 0.154        & 0.276        \\
\multicolumn{1}{c|}{}                         & 24  & {\color{red}\textbf{0.095}} & {\color{red}\textbf{0.188}} & {\color{blue}\underline{0.096}}    & {\color{blue}\underline{0.196}} & 0.126         & 0.236         & 0.115          & 0.218          & 0.117           & 0.226       & 0.137          & 0.246          & 0.115           & 0.219           & 0.224         & 0.281         & 0.141         & 0.238         & 0.248        & 0.353        \\
\multicolumn{1}{c|}{}                         & 48  & {\color{red}\textbf{0.140}} & {\color{red}\textbf{0.222}} & {\color{blue}\underline{0.149}}    & 0.244       & 0.223         & 0.320         & 0.192          & 0.294          & 0.169           & 0.268       & 0.265          & 0.343          & 0.186           & {\color{blue}\underline{0.235}}     & 0.321         & 0.354         & 0.198         & 0.283         & 0.440        & 0.470        \\
\multicolumn{1}{c|}{}                         & 96  & {\color{blue}\underline{0.239}}    & {\color{red}\textbf{0.267}} & 0.253          & 0.309       & 0.284         & 0.315         & 0.346          & 0.390          & 0.233           & 0.306       & 0.410          & 0.407          & {\color{red}\textbf{0.221}}  & {\color{red}\textbf{0.267}}  & 0.408         & 0.417         & 0.320         & 0.351         & 0.674        & 0.565        \\ 
\cmidrule(lr){2-22}
\multicolumn{1}{c|}{}                         & Avg & {\color{red}\textbf{0.137}} & {\color{red}\textbf{0.211}} & {\color{blue}\underline{0.142}}    & 0.229       & 0.180         & 0.266         & 0.183          & 0.271          & 0.150           & 0.246       & 0.226          & 0.299          & 0.150           & {\color{blue}\underline{0.226}}     & 0.280         & 0.321         & 0.193         & 0.271         & 0.379        & 0.416        \\ 
\specialrule{0.10em}{1pt}{1pt}
\multicolumn{2}{c|}{$1^{\text{st}}$ Count}         & {\color{red}\bf{38}}         & \multicolumn{1}{c|}{\color{red}\bf{57}}         &  {\color{blue}\underline{13}}     & \multicolumn{1}{c|}{0}      & 5       & \multicolumn{1}{c|}{0}      & 1      & \multicolumn{1}{c|}{1}      & 2      & \multicolumn{1}{c|}{0}      & 3      & \multicolumn{1}{c|}{2}  & 6   &\multicolumn{1}{c|}{1}      &  0    & \multicolumn{1}{c|}{0}   &  0     & \multicolumn{1}{c|}{0}    & 0    & 0    \\
\specialrule{0.15em}{2pt}{0pt}
\end{tabular}}
\end{table*}

\section{Discussion}

Despite its compelling performance, PAMNet exhibits certain limitations that present opportunities for future research. Two primary constraints are identified and discussed below, followed by corresponding future exploration directions.

\subsection{Potential limitations}

\textbf{Dependency on a Predefined Cycle Length $c$}: PAMNet's cyclical conditioning mechanism relies on a pre-specified cycle length $c$ (e.g., 24 for hourly daily cycles). 
While empirically robust to minor misspecification, this design inherently assumes a single, dominant, and stationary periodicity is known a priori. 
In real-world scenarios, time series often exhibit complex, multi-scale periodicities (e.g., combined daily, weekly, and annual cycles) or experience period shifts due to external events (e.g., daylight saving time, holiday schedules). 
A rigid $c$ may limit the model's adaptability to such non-stationary or composite cyclical patterns, potentially constraining its generalization in domains where underlying periods are unclear or variable.

\textbf{Lack of Explicit Cross-Frequency Interaction Modeling}: The core innovation of PAMNet lies in the decoupled modulation of phase and amplitude within a single presumed periodic component. However, it does not explicitly model the interactions between different periodic scales (frequencies). For instance, the amplitude of a daily pattern may be modulated by the day of the week (a weekly cycle), a phenomenon known as amplitude-amplitude coupling across frequencies. By treating periodicity through a single-scale lens, PAMNet may not fully capture these hierarchical or nested cyclical structures that are prevalent in complex systems, potentially leaving predictive information unexploited.

\subsection{Future Exploration Directions}

\textbf{Addressing Cycle Length Dependency}:
Future work should aim to reduce the model's reliance on a fixed, pre-defined $\bf{c}$. 
Promising directions include:
\begin{compactitem}
    \item Developing adaptive period detection modules that can learn or infer dominant cycle lengths directly from the data, possibly via learnable frequency embeddings or spectral analysis layers integrated into the training process.
    \item Exploring multi-cycle conditioning, where the model employs a set of basis cycle lengths (e.g., for daily, weekly, seasonal patterns) and learns to attend to or combine them dynamically based on the input context.
    \item Investigating soft or hierarchical period representations that can capture vague or time-varying periodicities without committing to a single integer cycle length.
\end{compactitem} 

\textbf{Enabling Cross-Frequency Interaction Modeling}:
To capture richer cyclical dynamics, extending PAMNet's modulation paradigm to multiple interacting frequencies is a natural next step. 
Research could explore: 
\begin{compactitem}
    \item Multi-branch modulation architectures with dedicated phase-amplitude modulators for different periodic bases, coupled with a fusion mechanism (e.g., cross-attention) to model their interactions.
    \item Frequency-domain modulation networks that operate on spectral representations of the time series, allowing explicit modeling of cross-frequency couplings in a more natural space.
    \item Hierarchical modulation schemes where the modulated output of a longer-period cycle (e.g., weekly) serves as a conditioning input or carrier for a shorter-period cycle (e.g., daily), explicitly modeling the nested structure of real-world rhythms.
\end{compactitem}

Addressing these limitations would enhance PAMNet's flexibility and expressive power for modeling the intricate, multi-scale periodicities inherent in real-world time series, further solidifying the phase-amplitude decoupling paradigm as a foundational approach for cyclical pattern modeling.


\end{document}